%% file: main.tex
\documentclass[manuscript, screen]{acmart}
\acmJournal{TOSEM}
\AtBeginDocument{%
  }

\setcopyright{none}

\input{macros}
\begin{document}

\title{An Empirical Study of Reasoning Steps in Thinking Code LLMs}


 \author{Haoran Xue}
 \email{hrx00@yorku.ca}
 \orcid{0009-0005-1937-3733} 
 \affiliation{%
   \institution{York University}
   \streetaddress{4700 Keele St.}
   \city{North York}
   \state{Ontario}
   \country{Canada}
   \postcode{M3J 1P3}
 } 

 \author{Gias Uddin}
 \email{guddin@yorku.ca}
 \affiliation{%
   \institution{York University}
   \streetaddress{4700 Keele St.}
   \city{North York}
   \state{Ontario}
   \country{Canada}
   \postcode{M3J 1P3}
 } 

 \author{Song Wang}
 \email{wangsong@yorku.ca}
 \affiliation{%
   \institution{York University}
   \streetaddress{4700 Keele St.}
   \city{North York}
   \state{Ontario}
   \country{Canada}
   \postcode{M3J 1P3}
 } 








\renewcommand{\shortauthors}{Xue et al.}

\input{sec/abstract}



\maketitle

\input{sec/intro}
\input{sec/studysetup}
\input{sec/resultsReasoningProcess}

\input{sec/resultsanalysis}

\input{sec/discussion}

\input{sec/relatedwork}
\input{sec/threads}
\input{sec/conclusion}

\bibliographystyle{ACM-Reference-Format}
\bibliography{paper}

\appendix









\end{document}

%% file: macros.tex
\usepackage{soul}
 \usepackage{url}

\usepackage{graphicx}
\usepackage[export]{adjustbox}
\usepackage{xcolor}
\usepackage{enumitem}
\usepackage{booktabs}
\usepackage{array}
\usepackage{tabularx}
\usepackage{multirow}
\usepackage{subcaption}
\usepackage{tikz}
\usetikzlibrary{calc,positioning,arrows.meta,patterns}
\usepackage{pgfplots}
\pgfplotsset{compat=1.16}
\usepgfplotslibrary{groupplots}
\definecolor{Gemini25LightBlue}{RGB}{153,204,255}
\definecolor{O3MiniPink}{RGB}{255,153,204}
\definecolor{Gemini20Beige}{RGB}{230,220,200}
\definecolor{DeepseekGray}{RGB}{150,150,150}
\definecolor{QwenLightPurple}{RGB}{200,170,255}
\definecolor{ClaudeLightGreen}{RGB}{180,220,180}

\usepackage[most]{tcolorbox}
\tcbuselibrary{theorems}

\makeatletter
\AtBeginDocument{%
  \@ifpackageloaded{hyperref}{%
    \pdfstringdefDisableCommands{%
      \def\textcolor#1#2{#2}%
      \def\textbf#1{#1}%
      \def\textit#1{#1}%
    }%
  }{}%
}
\makeatother

\newtcbtheorem[auto counter, number within=section]{observation}{Observation}%
{%
  enhanced,
  breakable,
  boxrule=0.5pt,
  colback=gray!10,
  colframe=black!60,
  arc=4pt,
  left=6pt,
  right=6pt,
  top=6pt,
  bottom=6pt,
  boxsep=0pt,
  fonttitle=\bfseries,
  title={Observation~\thetcbcounter.},%
  phantom={\hypertarget{tcb@cnt@observation.\thetcbcounter}{}},%
  label type=observation%
}{obs}
\makeatletter
\@ifpackageloaded{hyperref}{\hypersetup{hidelinks}}{}
\makeatother

\makeatletter
\AtBeginDocument{%
  \@ifpackageloaded{hyperref}{%
    \pdfstringdefDisableCommands{%
      \def\textcolor#1#2{#2}%
    }%
  }{}%
}
\makeatother

\definecolor{songcolor}{RGB}{191,191,191}



%% file: sec/abstract.tex
\begin{abstract}

Thinking Large Language Models (LLMs) generate explicit intermediate reasoning traces before final answers, potentially improving transparency, interpretability, and solution accuracy for code generation. However, the quality of these reasoning chains remains underexplored. We present a comprehensive empirical study examining the reasoning process and quality of thinking LLMs for code generation. We evaluate six state-of-the-art reasoning LLMs (DeepSeek-R1, OpenAI-o3-mini, Claude-3.7-Sonnet-Thinking, Gemini-2.0-Flash-Thinking, Gemini-2.5-Flash, and Qwen-QwQ) across 100 code generation tasks of varying difficulty from BigCodeBench. We quantify reasoning-chain structure through step counts and verbosity, conduct controlled step-budget adjustments, and perform a 21-participant human evaluation across three dimensions: efficiency, logical correctness, and completeness. Our step-count interventions reveal that targeted step increases can improve resolution rates for certain models/tasks, while modest reductions often preserve success on standard tasks, rarely on hard ones. Through systematic analysis, we develop a reasoning-problematic taxonomy, identifying completeness as dominant failure mode. Task complexity significantly impacts reasoning quality; hard problems are substantially more prone to incompleteness than standard tasks. Our stability analysis demonstrates that thinking LLMs maintain consistent logical structures across computational effort levels and can self-correct previous errors. This study provides new insights into the strengths and limitations of current thinking LLMs in software engineering. 
\end{abstract}

%% file: sec/intro.tex
\section{Introduction}
\label{sec:intro}

LLMs have achieved remarkable progress in software engineering tasks such as code generation~\cite{mu2024clarifygpt,jiang2024survey}, program translation~\cite{ibrahimzada2024alphatrans,luo2024repoagent,bairi2024codeplan}, and bug repair~\cite{fan2023automated,xia2023automated}. Despite achieving strong pass@1 accuracy on many benchmarks \cite{chen2021evaluatinglargelanguagemodels, austin2021programsynthesislargelanguage}, conventional (non-thinking) LLMs face structural limitations in complex coding tasks. 
First, their reasoning is not effectively transparent: even when correct code is produced, the intermediate hypotheses and decision points remain latent, preventing independent verification of the solution path and inhibiting process-level auditing \cite{turpin2023language}. Prompted rationales such as chain-of-thought can be elicited, but they are not guaranteed to be faithful to the model’s internal computation \cite{turpin2023language}. Second, these models exhibit fragility on multi-step programming tasks that demand principled decomposition (from informal requirements to subproblems), systematic edge-case coverage, and the coordination of multiple concepts (APIs, state, concurrency, resource constraints); performance often depends sensitively on prompts and decoding heuristics rather than robust, controllable planning~\cite{lightman2023letsverifystepstep}. Third, when generation fails, developers receive little process signal, i.e., only an incorrect final artifact, making it difficult to localize the breakdown (e.g., misread specification vs flawed algorithmic plan vs implementation slip), which slows debugging and hinders iterative improvement \cite{liao2023ai}. 

Recent thinking LLMs are models that explicitly externalize intermediate reasoning traces before producing final solutions (Fig. \ref{fig:non-thinking vs. thinking}). These advanced LLMs represent a fundamental architectural evolution from the black-box reasoning of traditional models to transparent, step-by-step cognitive processes. Thinking LLMs, exemplified by models such as OpenAI's o1 and o3 series \cite{openai2024openaio1card, openai_o3_mini}, DeepSeek-R1 \cite{marjanovi2025deepseekr1thoughtologyletsthink}, and Claude's reasoning variants \cite{anthropic2025claude37}, generate explicit chains of thought that articulate their problem understanding, solution strategy, implementation approach, and consideration of constraints and edge cases. By articulating their reasoning, these models are expected not only to generate correct code but also to provide explanations that align with program semantics, thereby improving trustworthiness and robustness in practical use~\cite{kojima2022large}.

\input{fig/comparison}

Despite these promises, the reliability and robustness of LLM-generated reasoning remain underexplored. Prior work on evaluating LLMs for code tasks has largely focused on the correctness of final outputs, e.g., pass@k for code generation \cite{chen2021evaluatinglargelanguagemodels, austin2021programsynthesislargelanguage, jain2024livecodebenchholisticcontaminationfree}.
While these studies reveal performance differences across different LLM models, they offer little insight into the quality of the reasoning process itself. As a result, we analyzed the step count and verbosity of the reasoning chains by different models, and then we ask in this paper: \textbf{How good is the reasoning quality produced by thinking LLMs as perceived by humans while using the LLMs for coding tasks?} To answer this question, we have conducted an empirical study to assess reasoning quality in thinking LLMs for code generation through the lenses of developers as human participants in our study.

We frame our study around two research questions. \textbf{RQ1 (Reasoning Process Analysis)} examines how thinking LLMs structure their chains. We explore this question from three sub-questions. First, we investigate how many steps thinking LLMs typically produce per task and how this differs between successful versus failed cases and between \textit{Hard} versus \textit{Full} tasks (RQ1.1). Second, we investigate whether changes in step count have a causal effect on resolution rates, specifically whether increasing step count improves a model's resolution rate (RQ1.2.1) and whether reducing step count degrades it (RQ1.2.2). Third, we examine whether verbosity in the reasoning chain is associated with resolution success (RQ1.3). \textbf{RQ2 (Reasoning Quality Analysis)} investigates the reasoning contents that identify patterns that characterize problematic reasoning (RQ2.1), and evaluates reasoning traces through three criteria—efficiency (RQ2.2), logical consistency (RQ2.3), and completeness (RQ2.4). We also examine how task complexity (\textit{Hard} vs. \textit{Full}) affects both reasoning quality and solution correctness across models (RQ2.5).

In our study, we quantify the reasoning chains by the six state-of-the-art reasoning LLMs:  DeepSeek-R1, OpenAI-o3-mini, Claude-3.7-Sonnet-Thinking, Gemini-2.0-Flash-Thinking, Gemini-2.5-Flash, and Qwen-QwQ, via step count and verbosity. We record the average steps produced by the thinking models on the success tasks and failed tasks, and to further explore the effect of step counts on the resolution rate changes of solving the code generation tasks, we run two controlled interventions experiments on the six reasoning LLMs: (1) \textit{think-deeper} by prompting the models to "think deeper" in expanding their reasoning chains uniformly increase from the original step count (from 10\% to 100\%) on prior failures to measure the resolution rate changes; and (2) \textit{reduce-step} by reducing steps progressively in the models' reasoning chains on prior successes to explore the maximum survival space. To assess the reasoning quality from the developers' perspectives, we conduct a human evaluation study in which participants evaluate the six reasoning LLMs on 100 diverse code generation tasks of varying difficulty levels drawn from BigCodeBench~\cite{zhuo2024bigcodebenchbenchmarkingcodegeneration}. Unlike prior work that only measures the correctness of final code outputs, we ask participants to analyze the reasoning traces themselves to answer important questions about how different LLMs ``think'' when approaching programming problems, potentially identifying strengths and weaknesses in their reasoning patterns. A total of 21 participants were recruited to label the reasoning steps generated in 600 coding tasks. They evaluated the reasoning traces along three key dimensions, i.e., efficiency~\cite{xia2025evaluating}, logic consistency~\cite{lee2025evaluatingstepbystepreasoningtraces}, and completeness \cite{brassard2024acorn}. We also experimented on OpenAI's o3-mini to provide a stability discussion on thinking LLMs, which involves setting the model with different thinking efforts due to its supportiveness. This allows us to measure stability by computing content similarity and changes in the resolution rate. In addition, we discuss the capability of self-correctness in thinking LLMs by guiding failures in the prompt to see if the model can fix them. 

Through our analysis, we found that some models like Gemini-2.0-FT may benefit from longer chains to get higher success rate on \textit{Hard} problems (Spearman value $\rho = 0.79$), but steps show less effect on the success rate on general tasks. Targeted step increases can occasionally improve resolution rates for certain models, but the relationship is non-monotonic and model-specific. Modest reductions of 10-30\% in reasoning steps often preserve success rates in standard tasks but significantly degrade performance on hard problems. To further identify the problematic patterns in the reasoning chains, we develop a novel taxonomy across the three metrics we proposed, identifying completeness issues (44.5\%), particularly the lack of edge case handling (32.17\%), as the most critical and frequent problems. We also want to know the impact of task complexity on reasoning quality, and we found that completeness showed stronger correlations with failure rates on hard problems (Spearman value $\rho = -0.219$) compared to standard tasks (Spearman value $\rho = -0.096$). Our stability discussion demonstrates that thinking LLMs (o3-mini) maintain consistent logical structures across different computational effort levels, and the model can self-correct errors in previous runs with and without guidance. 

This paper makes the following contributions:

\begin{itemize}[leftmargin=10pt]
    \item \textbf{Comprehensive evaluation of reasoning quality.} We systematically evaluate six state-of-the-art reasoning Large Language Models (LLMs) on 100 code generation tasks across hard and standard difficulties, analyzing step and verbosity counts and adjusting step count budgets to measure the resolution rate changes. 

    \item \textbf{Human-centered assessment of reasoning.} We conduct a 21-participant human evaluation work to evaluate reasoning quality along efficiency, logic consistency, and completeness criteria, providing insights into how humans perceive LLM reasoning.

    \item \textbf{Taxonomy of problematic reasoning.} We analyze the problematic reasoning traces annotated by participants and develop a taxonomy of reasoning problematic patterns, characterizing common problematic patterns and their frequencies. 
    
    \item \textbf{Analysis of stability and adaptability.} We investigate the stability of thinking Large Language Models (LLMs) by examining them with different reasoning effort levels and adaptability by analyzing whether and how they can refine or self-correct previously flawed reasoning when guided or unguided, re-prompted, thereby revealing their potential for iterative improvement. 

\end{itemize}

\textbf{Availability:} We release the data and source code of our experiments
to enable other researchers to replicate and extend our study https://github.com/xhinini/Reasoning-LLMs/tree/main.

%% file: fig/comparison.tex
\begin{figure}[H]
  \centering
  \includegraphics[width=\linewidth,    trim={0.1cm 17cm 0 0.1cm 2.5cm}, clip]{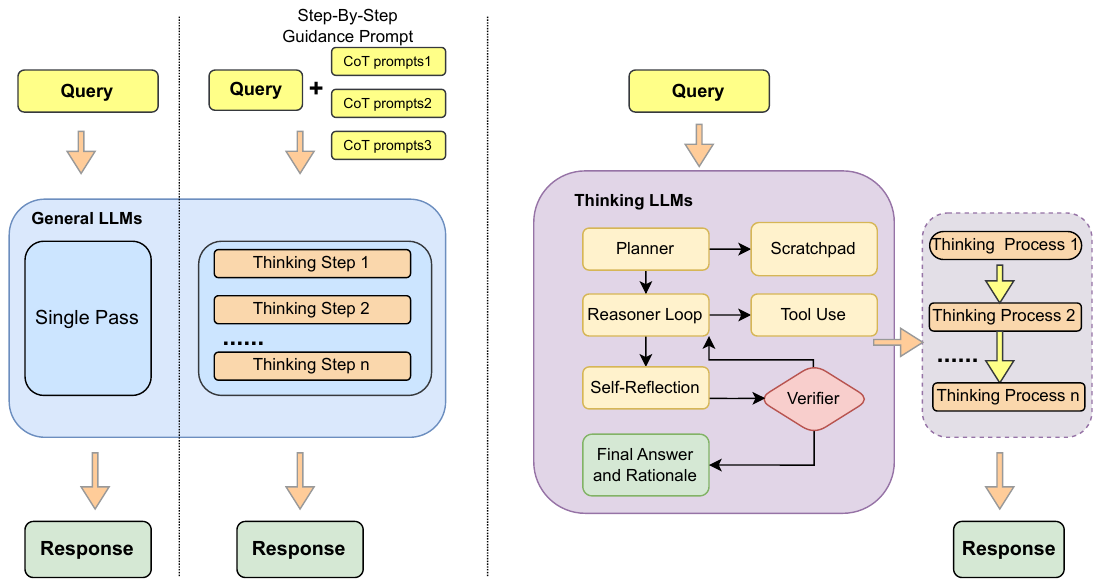}
  \caption{Processing flow comparison: Non-thinking vs. LLMs with CoT vs. Thinking LLMs}
  \label{fig:non-thinking vs. thinking}
\end{figure}

%% file: sec/studysetup.tex
\section{Study Setup}
\label{sec:empirical}

\input{fig/overall_bothrq}
This section describes the study setup for our study of the two RQs, including the studied tasks, studied thinking LLMs, and data collection, the whole workflow as presented in Fig. \ref{fig:overview_study}.

\subsection{Studied Code Generation Tasks}
\label{sec:code-generation-tasks}
We selected \textit{BigCodeBench} benchmark as our evaluation foundation due to its comprehensive coverage of real-world programming scenarios~\cite{zhuo2024bigcodebenchbenchmarkingcodegeneration}. \textit{BigCodeBench} is a large-scale benchmark containing 1140 programming tasks that span 723 function calls across 139 popular Python libraries in seven domains: computation, visualization, general utilities, time operations, system operations, networking, and cryptography \cite{zhuo2024bigcodebenchbenchmarkingcodegeneration}. Unlike traditional coding benchmarks such as \textit{HumanEval} \cite{chen2021evaluatinglargelanguagemodels} and \textit{MBPP} \cite{austin2021programsynthesislargelanguage} that focus on short, self-contained algorithmic problems, \textit{BigCodeBench} emphasizes practical programming tasks requiring complex instruction following and multi-library integration. 

Given the intensive nature of human evaluation required for assessing reasoning quality, we constructed a statistically representative subset from the full benchmark. From the total 1140 programming tasks, we randomly selected 100 tasks using standard statistical sampling methodology with a 95\% confidence level and a 5\% margin of error. This sampling process is standard in statistical analyses and widely used in software engineering empirical studies \cite{singh2014sampling}. In addition, to ensure our sample preserves the benchmark's structural characteristics, we employed proportional random sampling that maintains the original difficulty distribution: The full benchmark contains 148 ``Hard'' tasks (13.16\%) which are filtered based on three difficulty criteria: requiring more than two libraries, having solutions longer than the average length (426 tokens), and achieving solve rates below 50\% across evaluated models \cite{zhuo2024bigcodebenchbenchmarkingcodegeneration}and 992 remaining ``Full'' tasks (87\%), 
while our sample proportionally includes 14 tasks from the Hard set (14\%) and 86 tasks from the Full set (86\%), ensuring balanced coverage across complexity levels. 

The benchmark provides two evaluation splits, i.e., \textit{BigCodeBench-Complete}, which uses structured docstrings with comprehensive technical details, and \textit{BigCodeBench-Instruct}, which presents tasks as concise, natural-language instructions closer to how users typically express requirements \cite{zhuo2024bigcodebenchbenchmarkingcodegeneration}. For our study, we focus on the \textit{BigCodeBench-Instruct} split, which aligns with our goal of evaluating reasoning in realistic scenarios where models are used.

\subsection{Studied Reasoning LLMs}

This study employs a diverse collection of six thinking LLMs with varying sizes, versions, and release dates (Table \ref{tab:studied-llms}). Our selection includes both open-source and closed-source models. We accessed the models through their official APIs.

\textbf{DeepSeek-R1}: represents a groundbreaking advancement in reasoning LLMs \cite{marjanovi2025deepseekr1thoughtologyletsthink}. Built upon the DeepSeek-V3 base model with 671B total and 37B active parameters \cite{deepseekai2025deepseekv3technicalreport}, DeepSeek-R1 was developed through a sophisticated four-stage training pipeline involving reinforcement learning using Group Relative Policy Optimization (GRPO) \cite{deepseekai2025deepseekr1incentivizingreasoningcapability}. 

\textbf{OpenAI-o3-mini}: represents a significant advancement in cost-effective AI reasoning \cite{openai_o3_mini}. This powerful, compact model delivers exceptional capabilities in various domains, particularly excelling in mathematics and coding, while maintaining the low cost and reduced latency that make it highly accessible for widespread use \cite{openai_o3_mini}. 

\textbf{Claude 3.7 Sonnet-thinking}: is an advanced and intelligent model that excels in reasoning, complex problem-solving, coding, and instruction-following tasks \cite{anthropic2025claude37}. 

\textbf{Gemini-2.0-Flash-Thinking}: is an experimental model developed by Google's team, representing an experimental advancement in reasoning-capable language models \cite{geminiteam2025geminifamilyhighlycapable}. It demonstrates enhanced reasoning capabilities compared to its predecessor, Gemini-2.0-Flash \cite{qiu2025quantifyingreasoningabilitiesllms}. 

\textbf{Gemini-2.5-Flash}: represents a significant advancement in Large Language Models (LLMs) architecture, specifically excels in complex reasoning, mathematical computation, and coding \cite{google_gemini_models_2025}. According to Google, the ``thinking'' capabilities of the model enable enhanced accuracy and sophisticated contextual understanding through explicit reasoning processes \cite{google_gemini_models_2025}. 

\textbf{Qwen-QwQ}: is developed by the Qwen team, is an experimental model with 32 billion parameters, representing a significant advancement in open-source reasoning-capable language models, featuring its thinking and reasoning capabilities \cite{zheng2024processbench}. The model shows significantly enhanced performance on complex problems \cite{zheng2024processbench}. 

\input{tab/studied_llms}

\subsection{Data Collection} 

To enable systematic evaluation of reasoning quality across different LLMs, we conducted a comprehensive data collection process involving six reasoning-capable models. This section describes the process by which we generated, processed, and organized the reasoning data for manual evaluation. 

We first ran six different reasoning LLMs to examine their reasoning content and step-by-step processes for each code generation task. This stage focused on developing systematic methods to analyze how LLMs ``think'' when approaching code generation. Notably, models vary in how they present their reasoning traces. For example, DeepSeek-R1 provides a built-in separation between the reasoning process and the final response, offering distinct fields, i.e., ``content'' and ``reasoning\_content'', that explicitly distinguish between internal reasoning and direct solutions~\cite{deepseek-api}. 
This architectural feature enables direct access to the raw reasoning trace via the ``reasoning\_content'' field. In contrast, models such as Gemini-2.0-Flash-Thinking and Qwen-QwQ do not make this distinction, instead embedding the reasoning process and final answer together in a single output. A recent study reported a similar observation~\cite{qiu2025quantifyingreasoningabilitiesllms}. Moreover, the raw reasoning traces are often lengthy and difficult to interpret, which presents challenges for human evaluation. To address this, we implemented a structured output format inspired by prior work, which guides LLMs to present their reasoning more systematically. Specifically, we prompted models to organize their reasoning into clearly delineated steps (e.g., <step1>, <step2>, etc.), with each step containing a specific reasoning action or decision. To ensure fair comparison across models, we applied this structured prompting approach uniformly, including to the DeepSeek-R1 model despite its built-in reasoning separation capabilities. We use this standardization to ensure consistent evaluation conditions across all evaluated models.

The data collection process proceeded in three phases. 
First, we executed each of the six reasoning models on our 100-task sample, storing all outputs, including both reasoning content and final solutions, in structured JSON files for systematic processing. Second, to facilitate human evaluation, we reformatted this data into human-readable text files, with each file containing exactly 10 tasks and the corresponding reasoning contents from all six models (60 reasoning sequences per file). Within each file, we organized content by task: presenting the problem description, followed by clearly labeled reasoning traces from each model (model1\_reasoning\_contents, model2\_reasoning\_contents, etc.). Finally, to ensure evaluation reliability and mitigate individual annotator bias, we created three identical copies of all materials, enabling independent assessment by three evaluators per task.

\input{tab/task_sum}
Table \ref{tab:reasoning_data} summarizes the reasoning data collected for manual evaluation, illustrating the substantial effort required for comprehensive assessment. Across the six models, we collected a total of 3,772 reasoning steps from 600 task instances. The average number of reasoning steps per task varied across models, ranging from 5.02 steps (Qwen-QwQ) to 8.2 steps (Gemini-2.0-Flash-Thinking), with an overall average of 6.29 steps per task. 

%% file: fig/overall_bothrq.tex
\begin{figure}[H]
  \centering
  \includegraphics[
    width=\linewidth,
    trim={1.5cm 19cm 5.5cm 4cm}, clip
  ]{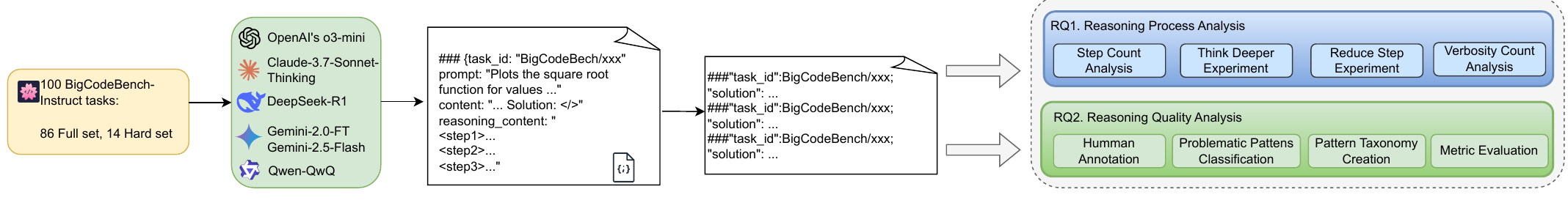}
  \caption{Overview of the Study Setup}
  \label{fig:overview_study}
\end{figure}

%% file: tab/studied_llms.tex


\begin{table}[t!] 
  \caption{Studied reasoning (``thinking'') LLMs. Dates/params shown only when verified; otherwise ``—''.}
  \label{tab:studied-llms}
  \small
  \setlength{\tabcolsep}{6pt}
  \renewcommand{\arraystretch}{1.12}
  \begin{tabularx}{\linewidth}{@{} >{\raggedright\arraybackslash}X c l l l @{}}
    \toprule
    \textbf{Model} & \textbf{Date} & \textbf{Provider} & \textbf{Params} & \textbf{Ref} \\
    \midrule
    DeepSeek\texttt{-}R1 & 2025.05 & DeepSeek & 671B & \cite{marjanovi2025deepseekr1thoughtologyletsthink, deepseekai2025deepseekv3technicalreport, deepseekai2025deepseekr1incentivizingreasoningcapability} \\
    OpenAI o3-mini & 2025.01 & OpenAI & — & \cite{openai_o3_mini} \\
    Claude 3.7 Sonnet-thinking & 2025.02 & Anthropic & — & \cite{anthropic2025claude37} \\
    Gemini-2.0-Flash-Thinking & 2025.01 & Google & — & \cite{geminiteam2025geminifamilyhighlycapable} \\
    Gemini-2.5-Flash & 2025.05 & Google & — & \cite{google_gemini_models_2025} \\
    Qwen-QwQ & 2024.11 & Qwen (Alibaba) & 32B & \cite{zheng2024processbench} \\
    \bottomrule
  \end{tabularx}
\end{table}

%% file: tab/task_sum.tex
\begin{table}[t!]
\centering
\caption{Summary of reasoning data collected for manual evaluation}
\label{tab:reasoning_data}
\begin{tabular}{lccc}
\toprule
\textbf{LLM Model} & \textbf{Total Tasks} & \textbf{Total Reasoning Steps} & \textbf{Average Steps per Task} \\
\midrule
OpenAI-O3-mini & 100 & 585 & 5.85 \\
Claude-3.7-Sonnet-Thinking & 100 & 570 & 5.7 \\
Qwen-QwQ & 100 & 502 & 5.02 \\
DeepSeek-R1 & 100 & 558 & 5.58 \\
Gemini-2.5-Flash & 100 & 737 & 7.37 \\
Gemini-2.0-Flash-Thinking & 100 & 820 & 8.2 \\
\midrule
\textbf{Total} & \textbf{600} & \textbf{3,772} & \textbf{6.29} \\
\bottomrule
\end{tabular}
\end{table}

%% file: sec/resultsReasoningProcess.tex
\section{Assessing the Reasoning Process in Thinking Code LLMs (RQ1)}
Prior work on LLM reasoning mainly centers on output correctness, leaving the structure and quality of intermediate reasoning largely unexamined \cite{yang2025code}. Understanding the quality of reasoning is crucial for building trust in LLM-generated solutions and identifying systematic weaknesses that prevent these models from solving complex programming tasks. However, fundamental questions about the structure of these reasoning processes remain unanswered. We do not yet know whether models' elaborate reasoning chains, characterized by more steps or greater verbosity, correlate with higher success rates or whether conciseness is more effective. Different models may employ distinct reasoning strategies: some might generate lengthy, detailed explorations, while others produce compact and direct solutions. Without a systematic analysis of these patterns, we cannot determine which approaches are most effective or whether reasoning length serves as a reliable indicator of solution quality. 

To address this gap and provide a foundational understanding of how ``thinking'' LLMs structure their reasoning processes, we study two complementary aspects of reasoning chain characteristics and their effects. First, we examine the quantitative structure of reasoning through step counts, as the number of reasoning steps may reflect the complexity of a model's problem-solving approach and potentially correlate with success rates. Second, we analyze reasoning verbosity to see whether a more detailed articulation of reasoning improves solution quality. Beyond the above analysis, we also run two causal tests that directly change the step budget: we add steps to re-try previously failed cases (\textit{think-deeper}) and we remove steps from previously successful cases to see how much we can tighten the chain while still succeeding (\textit{reduce-steps}).

We formalize these as: 
\begin{itemize}
\item \textbf{RQ1.1} How many steps do thinking LLMs typically produce per task, and how does this differ between successful vs.\ failed cases and between Hard vs.\ Full tasks?

\item \textbf{RQ1.2} Do changes in the number of step counts impact the resolution rates?

\begin{itemize}
    \item RQ1.2.1 Can an increase on step count can increase the resolution rate of a model?
    \item RQ1.2.2 Can a reduction in step count reduce the resolution rate of a model?
\end{itemize}

\item \textbf{RQ1.3}  Is verbosity in thinking linked to resolution success?
\end{itemize}

\subsection{Experiment Approach}
To systematically investigate these questions, we run all six models on the same 100 tasks to ensure fair, task-controlled comparisons across models and difficulty splits (\textit{Hard} vs. \textit{Full}). For each model and split, we log (i) the average number of steps on successes vs.\ failures, (ii) the overall resolution rate, and (iii) a task-controlled association between step count and pass/fail using Spearman’s rank correlation. Let \(X_i\) denote the step count and \(Y_i\in\{0,1\}\) the outcome for task \(i\); we compute
\begin{equation} 
\rho_s \;=\; \mathrm{corr}\!\left(\mathrm{rank}(X_1,\ldots,X_n),\,\mathrm{rank}(Y_1,\ldots,Y_n)\right),
\end{equation} 
where positive values indicate that, on the same tasks, longer reasoning chains tend to co-vary with higher success rates.
In addition to step counts, we compute the per-step verbosity (average number of words per step) to capture the depth of explanation within each reasoning step. To establish causal relationships between chain length and success, we conduct pre-specified step-budget adjustments that vary only the permitted chain length while keeping prompts and all other decoding settings fixed: \textit{Think-deeper intervention}: We increase step budgets for previously failed attempts and re-run the tasks to measure whether additional reasoning steps can recover failures.
\textit{Reduce-steps intervention}: We progressively decrease step budgets for previously successful attempts to identify the minimum chain length that preserves success, effectively finding the "critical threshold" beyond which further reduction causes failure.

By comparing resolution rates before and after these interventions, we can quantify the causal impact of reasoning chain length on solution quality.

\subsection{RQ1.1 Step Count Analysis}
\label{sec: rq1.1}

\input{fig/step_ana}
\paragraph{Approach}
This sub-question addresses the fundamental structural characteristic of reasoning chains' length. By comparing step counts across successful and failed cases, we can determine whether more elaborate reasoning processes lead to better outcomes or whether failed attempts are characterized by either premature termination (i.e., too few steps) or excessive exploration (i.e., too many steps). Furthermore, comparing \textit{Hard} versus \textit{Full} tasks reveals whether models adaptively adjust their reasoning depth based on perceived problem complexity. Answering this question provides essential baseline data on reasoning structure and helps identify whether step count serves as a meaningful indicator of reasoning quality.

\paragraph{Results}
As shown in Figure \ref{fig:steps-resolution-bars}, on the \textit{Hard} set, Gemini-2.0-FT demonstrates the clearest stepwise reasoning benefit: successful attempts average 13.75 steps vs. 8.10 on failures, with a strong positive correlation between step count and success (Spearman value $=0.79$). This model appears to benefit from extended exploration, in that the more steps it takes, the more likely it is to find the solution, achieving a 28.57\% resolution rate on \textit{Hard} set problems. O3-mini also shows a relatively high correlation (Spearman value $\rho = 0.219$).  In contrast, DeepSeek-R1 (Spearman value $\rho=-0.579$) and Qwen-QwQ show the opposite pattern that succeeds with fewer steps (Spearman value $\rho=-0.52$). This model appears to benefit more from conciseness, as longer chains may indicate the model is struggling or going down unproductive paths. Other models, such as Claude-3.7-sonnet-thinking, show weak step-success correlations on \textit{Hard} tasks, suggesting their performance is less dependent on chain length. Notably, when tasks become easier on the \textit{Full} set, all step-success correlations collapse toward zero, and resolution rates improve across the board (29.1\% to 55.8\% compared to 14.3\% to 35.7\% on \textit{Hard}). Furthermore, the average step counts between successful and failed attempts converge on \textit{Full} tasks (differences of $\leq$1 step for most models), contrasting sharply with the divergent patterns on \textit{Hard} tasks. This suggests that step count primarily matters when problems challenge a model's fundamental capabilities, meaning that easy problems do not require extended reasoning. In contrast, the optimal chain length for \textit{Hard} problems appears to be model-specific.  

\begin{observation}{}
{}Step count demonstrates model-specific and task-difficulty-dependent relationships with success. On \textit{Hard} set, some models show positive correlation like Gemini-2.0-FT ($\rho = 0.79$) and O3-mini ($\rho = 0.22$); while other models like Deepseek-R1 ($\rho = -0.579$) and Qwen-QwQ ($\rho = -0.52$) show negative correlation between the step count and success rate. On \textit{Full} tasks, all correlations approach zero, indicating that step count becomes irrelevant for easier problems. The substantial gap in average steps between success and failure on \textit{Hard} tasks (e.g., 5.65-step difference for Gemini-2.0-FT) versus minimal differences on \textit{Full} tasks ($\leq$1 step) suggests reasoning depth matters primarily at the edge of model capabilities.
\end{observation}

\subsection{RQ1.2 Step Count Effect}

To test whether step counts causally impact resolution rates, we ran two complementary interventions that directly manipulate chain length while holding all other decoding settings fixed. 
\subsubsection{RQ1.2.1 Think Deeper}
\paragraph{Approach}

For each previously failed attempt, we prompt the models to think more deeply by producing a longer chain that extends the original reasoning chains by 10\% to 100\%, and then re-run the task. We then record the change in resolution rate at each budget. We applied this procedure to each model with the following sample sizes of failed cases: DeepSeek-R1 (n = 66), o3-mini (n = 47), Claude-3.7-Sonnet-Thinking (n = 72), Gemini-2.0-FT (n = 60), Gemini-2.5 (n = 67), and Qwen-QwQ (n = 61).

\paragraph{Results}

\input{fig/889-thinkdeeper}

In our \textit{think-deeper} sweep that 10\% to 100\% extensions of the failed chains, the curve, presented in Fig. \ref{fig:resolution_rate_full_hard} shows small bands where resolution improves, followed by a flat or lower performance at large extensions. This non-monotonic pattern appears on both splits: on \textit{Full} set, some models like Gemini-2.5-Flash and Gemini-2.0-FT exhibit mid-range bumps (around +50–60\%) with no sustained upward trend; while Deepseek-R1 shows a later bump (around +80\%); Claude-3.7-Sonnet-Thinking remains relatively flat, and O3-mini changes little. On the \textit{Hard} set, the curves are more volatile. Claude-3.7-Sonnet-Thinking shows a single pronounced peak (+70\%), O3-mini spikes once (+80\%), and Gemini-2.0-FT shows no gain across the sweep. This greater volatility on Hard is expected because we have fewer evaluable cases per percentage bin, so each change in outcomes moves the rate more, producing sharper peaks and dips. 
We provide an example of a resolved task by applying ``\textit{think-deeper}'' to a previous unresolved task, as shown in Fig. \ref{fig:think-deeper-889-raw}, where we prompted the LLM to increase the step count by 50\% (from 6 steps to 9 steps), which fixed the previously unresolved attempt by adding an explicit FileNotFoundError re-raise, restricting imputation to numeric-only columns, and inserting an early empty-CSV return, after which the model successfully resolved the task.

\begin{observation}{}
{}A small, targeted step increase can help to increase the resolution rate,  but benefits are not monotonic and depend on the model and the specific step-increase band.
\end{observation}

\input{fig/think_deeper_line}

\input{fig/1128-reduce-steps}

\input{fig/reduce_step}
\subsubsection{RQ1.2.2 Reduce Steps}

\paragraph{Approach} 
For each previously successful attempt, we progressively lower the step budget to find the largest relative step reduction that still preserves success. We report this maximum per task, quantized to 0.1 (10\%) increments, so each task falls into exactly one reduction bin (e.g., 0–10\%, 10–20\%, \dots). Percentages across bins are therefore not cumulative. As the reduce steps experiment uses all originally successful tasks as the baseline, the per-model sample sizes are \(n{=}28\) for Claude-3.7-Sonnet-Thinking, \(n{=}53\) for o3-mini, \(n{=}34\) for DeepSeek-R1, \(n{=}39\) for Qwen-QwQ, \(n{=}40\) for Gemini-2.0-FT, and \(n{=}33\) for Gemini-2.5-Flash.

\paragraph{Results}
We plotted survival curves \(S(x)\) as shown in Figure \ref{fig:stepcut_survival}: \(S(x)\) is the fraction of originally passed tasks that still pass after an \(x\%\) step cut, where every row sums to 1 and curves anchor at \(S(0){=}1\).  On the \textit{Full} split, stronger models retain noticeable safe room: Gemini-2.5-Flash decays most slowly through the 10--40\% region, Gemini-2.0-FT shows moderate tolerance, and o3-mini largely withstands only 10--20\% cuts; DeepSeek-R1, Qwen-QwQ, and Claude-3.7-Sonnet-Thinking drop earlier. On the \textit{Hard} split, most curves collapse by a 10\% reduction—Claude-3.7-Sonnet-Thinking, Gemini-2.5-Flash, and Qwen-QwQ show no safe room—while only Gemini-2.0-FT and o3-mini retain limited mass into the 20--40\% range. Read left-to-right, the vertical separation around 10--30\% on \emph{Full} quantifies each model’s shortening tolerance, reinforcing the main takeaway: small reduction (\(\approx\)10--30\%) often preserves success on \textit{Full}, but rarely on \textit{Hard}. We provide an example as shown in the Fig. \ref{fig:step-reduce-1128} where we prompt the LLM to decrease the step count by 42\% compared to the original resolved chain, the shortened reasoning then failed to consider reading/parsing the JSON, skipped writing the timestamped output and returning its absolute path, and thus the model did not successfully resolve the task.

\begin{observation}{}
{}Reasoning chains exhibit model- and difficulty-dependent compressibility. On \textit{Full} tasks, top performers (Gemini-2.5-Flash, Gemini-2.0-FT) tolerate 10--40\% step reductions. On \textit{Hard} tasks, most models collapse at 10\% reduction, with only Gemini-2.0-FT and o3-mini maintaining limited tolerance to 20--40\%. Excessive shortening causes critical operation omission rather than graceful degradation.
\end{observation}

\input{fig/verbosity_ana}

\subsection{RQ1.3 Verbosity Analysis}
\paragraph{Approach}
While step count measures the breadth of reasoning, verbosity captures its depth, which is the level of detail and explanation within each step. This distinction is crucial because two reasoning chains with identical step counts may differ dramatically in their explanatory content. By measuring the average words per step and correlating verbosity with success rates, we can determine whether more detailed articulation improves solution quality or whether concise reasoning is more effective. 
Using the same task-controlled setup, we measure verbosity as the average number of words per step for each attempt. For every model and difficulty split, we compare verbosity on successes vs. failures (as shown in Figure \ref{fig:verbosity-resolution-bars}) and compute Spearman's r between verbosity and pass/fail. 

\paragraph{Results}

The verbosity analysis reveals that the relationship between verbosity and resolution success is model-specific rather than universal, with different models exhibiting distinct preferences that should be understood alongside their step-count behaviors, as discussed in Section~\ref {sec: rq1.1}. 

Gemini-2.0-FT succeeds with concise individual steps but utilizes many of them; successful attempts average only 44.27 words per step, compared to 66.59 words on failures (Spearman value $\rho$ = -0.67). This is complementary to Section~\ref{sec: rq1.1}, where we showed that Gemini-2.0-FT uses more total steps when succeeding (13.75 vs. 8.10 steps, Spearman value $\rho$ = 0.79), indicating that the model breaks down problems into numerous steps while keeping each step focused and efficient. When the model fails, it produces fewer total steps that are individually more verbose, suggesting it becomes stuck elaborating on particular points rather than systematically progressing through the problem. Claude-3.7-Sonnet-Thinking exhibits a similar, though less pronounced, pattern (43.24 vs. 40.96 words per step), maintaining relatively consistent conciseness across outcomes. Qwen-QwQ shows the inverse pattern on \textit{Hard} tasks: successful attempts are more verbose at 42.44 words per step compared to 29.67 on failures (Spearman value $\rho$ = 0.71). Connecting to section~\ref{sec: rq1.1}, we found that Qwen-QwQ succeeds with fewer total steps on Hard tasks (4.5 vs. 6.2 steps, Spearman value $\rho$ = -0.52), meaning this model compensates for shorter reasoning chains by providing more detailed exploration within each step. When Qwen-QwQ fails on \textit{Hard} tasks, it produces both more steps and less verbose ones, suggesting it engages in shallow exploration rather than focused, deep reasoning. However, this pattern does not hold consistently across difficulty levels. On \textit{Full} tasks, Qwen-QwQ shows slightly more steps when succeeding (5.06 vs. 4.8) while maintaining higher verbosity (39.23 vs. 33.89 words per step). This indicates that Qwen-QwQ's reasoning strategy adapts to task difficulty; on challenging problems, it succeeds by concentrating detailed reasoning into fewer steps. DeepSeek-R1 and Gemini-2.5-Flash maintain high verbosity regardless of outcome, with both achieving only 14.29\% resolution rate on \textit{Hard} tasks. Their high word counts suggest they are over-elaborating without effectively advancing the solution. 

\begin{observation}{}
{}Verbosity and step count interact in model-specific ways to predict success. Effective models employ distinct strategies: some succeed through many concise steps (Gemini-2.0-FT), others through fewer but more detailed steps (Qwen-QwQ on \textit{Hard} tasks). These patterns can shift with task difficulty, and high verbosity without strategic reasoning (DeepSeek-R1, Gemini-2.5-Flash) does not guarantee success.
\end{observation}

%% file: fig/step_ana.tex

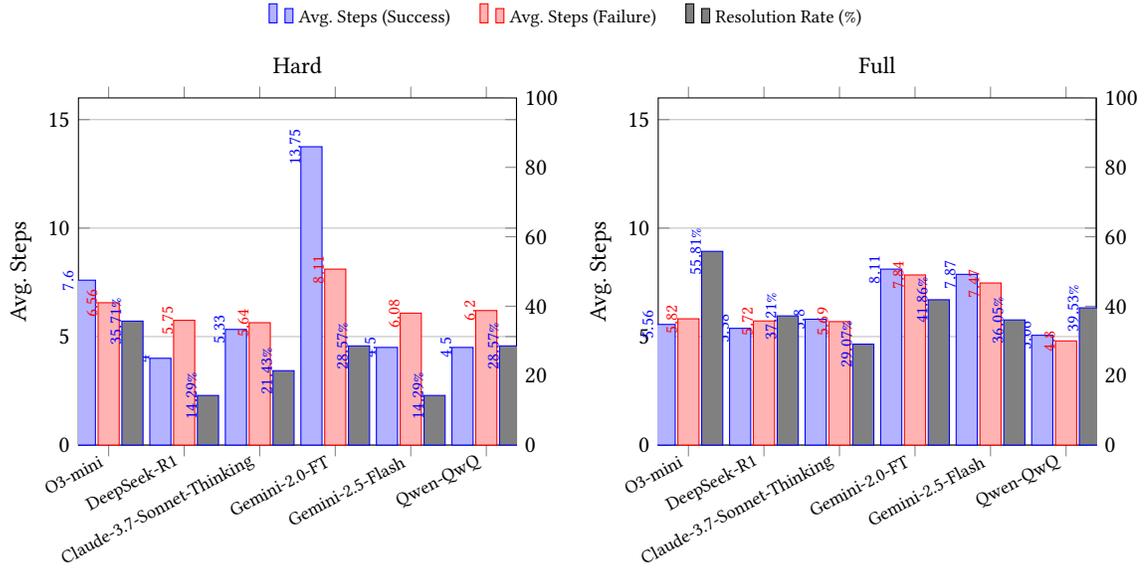
\begin{figure*}[t]
\centering
\pgfplotslegendfromname{stepsLegend}
\vspace{.4em}

\begin{subfigure}{0.49\textwidth}
\centering
\begin{tikzpicture}
\begin{axis}[
  ybar,
  bar width=8pt,
  width=\textwidth, height=6.2cm,
  ymin=0, ymax=16,                
  ytick={0,5,10,15},              
  ylabel={Avg. Steps},
  title={Hard},
  symbolic x coords={O3-mini,DeepSeek-R1,Claude-3.7-Sonnet-Thinking,Gemini-2.0-FT,Gemini-2.5-Flash,Qwen-QwQ},
  xtick=data,
  xticklabel style={rotate=28,anchor=east, font=\footnotesize},
  enlarge x limits=0.08,
  ymajorgrids,
  legend to name=stepsLegend,
  legend columns=3,
  legend style={/tikz/every even column/.append style={column sep=1em}, draw=none, font=\footnotesize},
  nodes near coords,
  point meta=y,
  every node near coord/.append style={
    font=\scriptsize, rotate=90, anchor=south, yshift=1pt,
    /pgf/number format/fixed, /pgf/number format/precision=2
  },
]
  \addplot+[ybar, bar shift=-9pt] coordinates
    {(O3-mini,7.6) (DeepSeek-R1,4) (Claude-3.7-Sonnet-Thinking,5.33)
     (Gemini-2.0-FT,13.75) (Gemini-2.5-Flash,4.5) (Qwen-QwQ,4.5)};
  \addlegendentry{Avg. Steps (Success)}

  \addplot+[ybar, bar shift=0pt] coordinates
    {(O3-mini,6.56) (DeepSeek-R1,5.75) (Claude-3.7-Sonnet-Thinking,5.64)
     (Gemini-2.0-FT,8.11) (Gemini-2.5-Flash,6.08) (Qwen-QwQ,6.2)};
  \addlegendentry{Avg. Steps (Failure)}

  \addlegendimage{ybar, fill=gray}
  \addlegendentry{Resolution Rate (\%)}
\end{axis}

\begin{axis}[
  ybar,
  bar width=8pt,
  width=\textwidth, height=6.2cm,
  ymin=0, ymax=100,
  axis y line*=right,
  axis x line=none,
  symbolic x coords={O3-mini,DeepSeek-R1,Claude-3.7-Sonnet-Thinking,Gemini-2.0-FT,Gemini-2.5-Flash,Qwen-QwQ},
  xtick=data,
  enlarge x limits=0.08,
  nodes near coords,
  point meta=y,
  nodes near coords={\pgfmathprintnumber{\pgfplotspointmeta}\%},
  every node near coord/.append style={
    font=\scriptsize, rotate=90, anchor=south, yshift=1pt,
    /pgf/number format/fixed, /pgf/number format/precision=2
  },
]
  \addplot+[ybar, bar shift=+9pt, fill=gray] coordinates
    {(O3-mini,35.71) (DeepSeek-R1,14.29) (Claude-3.7-Sonnet-Thinking,21.43)
     (Gemini-2.0-FT,28.57) (Gemini-2.5-Flash,14.29) (Qwen-QwQ,28.57)};
\end{axis}
\end{tikzpicture}
\end{subfigure}
\hfill
\begin{subfigure}{0.49\textwidth}
\centering
\begin{tikzpicture}
\begin{axis}[
  ybar,
  bar width=8pt,
  width=\textwidth, height=6.2cm,
  ymin=0, ymax=16,                
  ytick={0,5,10,15},              
  ylabel={Avg. Steps},
  title={Full},
  symbolic x coords={O3-mini,DeepSeek-R1,Claude-3.7-Sonnet-Thinking,Gemini-2.0-FT,Gemini-2.5-Flash,Qwen-QwQ},
  xtick=data,
  xticklabel style={rotate=28,anchor=east, font=\footnotesize},
  enlarge x limits=0.08,
  ymajorgrids,
  legend=false, 
  nodes near coords,
  point meta=y,
  every node near coord/.append style={
    font=\scriptsize, rotate=90, anchor=south, yshift=1pt,
    /pgf/number format/fixed, /pgf/number format/precision=2
  },
]
  \addplot+[ybar, bar shift=-9pt] coordinates
    {(O3-mini,5.56) (DeepSeek-R1,5.38) (Claude-3.7-Sonnet-Thinking,5.8)
     (Gemini-2.0-FT,8.11) (Gemini-2.5-Flash,7.87) (Qwen-QwQ,5.06)};

  \addplot+[ybar, bar shift=0pt] coordinates
    {(O3-mini,5.82) (DeepSeek-R1,5.72) (Claude-3.7-Sonnet-Thinking,5.69)
     (Gemini-2.0-FT,7.84) (Gemini-2.5-Flash,7.47) (Qwen-QwQ,4.8)};
\end{axis}

\begin{axis}[
  ybar,
  bar width=8pt,
  width=\textwidth, height=6.2cm,
  ymin=0, ymax=100,
  axis y line*=right,
  axis x line=none,
  symbolic x coords={O3-mini,DeepSeek-R1,Claude-3.7-Sonnet-Thinking,Gemini-2.0-FT,Gemini-2.5-Flash,Qwen-QwQ},
  xtick=data,
  enlarge x limits=0.08,
  nodes near coords,
  point meta=y,
  nodes near coords={\pgfmathprintnumber{\pgfplotspointmeta}\%},
  every node near coord/.append style={
    font=\scriptsize, rotate=90, anchor=south, yshift=1pt,
    /pgf/number format/fixed, /pgf/number format/precision=2
  },
]
  \addplot+[ybar, bar shift=+9pt, fill=gray] coordinates
    {(O3-mini,55.81) (DeepSeek-R1,37.21) (Claude-3.7-Sonnet-Thinking,29.07)
     (Gemini-2.0-FT,41.86) (Gemini-2.5-Flash,36.05) (Qwen-QwQ,39.53)};
\end{axis}
\end{tikzpicture}
\end{subfigure}

\caption{Average reasoning \emph{steps} (success/failure) and resolution rate for each model, split by difficulty. Both left y-axes share the same range and tick interval (0, 5, 10, 15). Resolution rate uses the right y-axis.}
\label{fig:steps-resolution-bars}
\end{figure*}

%% file: fig/889-thinkdeeper.tex
\begin{figure}[t]
  \centering
  \begin{tcolorbox}[
      enhanced,
      width=\linewidth,
      boxsep=2pt,left=3pt,right=2.9pt,top=3pt,bottom=3pt,
      colback=white,colframe=black!25,
      fontupper=\scriptsize,
      before skip=2pt,after skip=2pt
  ]
  
\begin{tcolorbox}[
    colback=gray!15,
    colframe=gray!15,
    boxsep=2pt,
    left=2pt,right=2pt,top=2pt,bottom=2pt,
    sharp corners,
    boxrule=0pt
]
\href{https://huggingface.co/datasets/bigcode/bigcodebench-hard/viewer}{\textbf{Problem (BigCodeBench/889)}}\quad
\\ 
Load CSV to \texttt{DataFrame}; if empty, return empty \texttt{DataFrame}; replace \texttt{NaN} in \emph{numeric} columns with the column mean; raise \texttt{FileNotFoundError} if missing. Must start with the given import block and signature.
\end{tcolorbox}

\vspace{2pt}
\hrule height 0.5pt
\vspace{3pt}

\begin{minipage}[t]{0.48\linewidth}
\raggedright
\textbf{Original: 6 steps, unresolved}\\[3pt]
\begin{enumerate}\setlength{\itemsep}{1pt}\setlength{\parskip}{0pt}\setlength{\topsep}{0pt}
\item The problem requires loading a CSV file into a pan\-das \texttt{DataFrame}. The function takes a directory path and a filename. The first step is to combine these to get the full path to the CSV file using \texttt{os.path.join}.
\item Load the CSV file into a pandas \texttt{DataFrame} us\-ing \texttt{pd.read\_csv()}. This function inherently raises \texttt{FileNotFoundError} if the file does not exist at the specified path, which matches one of the require\-ments.
\item Check if the loaded \texttt{DataFrame} is empty. If an empty CSV is passed, an empty \texttt{DataFrame} should be returned. If \texttt{df.empty} is \texttt{True}, return it immediately.
\item Replace \texttt{NaN} in numeric columns with the mean of their respective columns. Use \texttt{df.mean()} to compute means for numeric columns (returns a \texttt{Series}).
\item Use \texttt{df.fillna()} with the \texttt{Series} from \texttt{df.mean()} to fill \texttt{NaN}s per column. Either \texttt{inplace=True} or \texttt{df = df.fillna(df.mean())}.
\item Return the processed \texttt{DataFrame}.
\end{enumerate}
\end{minipage}%
\hfill\vline\hfill%
\begin{minipage}[t]{0.5\linewidth}
\raggedright
\textbf{Refined with think deeper: 9 steps +50\%, resolved}\\[3pt]
\begin{enumerate}\setlength{\itemsep}{1pt}\setlength{\parskip}{0pt}\setlength{\topsep}{0pt}
\item \colorbox{green!20}{Understand the problem: load a CSV into a pandas} \colorbox{green!20}{\texttt{DataFrame}, replace \texttt{NaN} in numeric columns with} \colorbox{green!20}{the mean, and return the modified \texttt{DataFrame}.} \colorbox{green!20}{Also handle \texttt{FileNotFoundError} and the empty CSV case.}
\item \colorbox{green!20}{Plan the approach: read CSV → identify numeric} \colorbox{green!20}{columns → compute per-column means → fill →} \colorbox{green!20}{return; include explicit error handling and empty-} \colorbox{green!20}{\texttt{DataFrame} handling.}
\item \colorbox{green!20}{Handle \texttt{FileNotFoundError}: implement a} \colorbox{green!20}{\texttt{try/except} to catch missing file; if caught,} \colorbox{green!20}{re-raise to satisfy the requirement.}
\item Load the CSV into a \texttt{DataFrame} using \texttt{pd.read\_csv()}.
\item Check for an empty \texttt{DataFrame}. If empty, return an empty \texttt{DataFrame}.
\item \colorbox{green!20}{Identify numeric columns using} \colorbox{green!20}{\texttt{df.select\_dtypes(include=np.number)}.}
\item Calculate the mean for each identified numeric col\-umn (e.g., \texttt{numeric.mean()} or \texttt{df.mean()} restricted to numeric).
\item \colorbox{green!20}{Replace \texttt{NaN} values: iterate through numeric} \colorbox{green!20}{columns and use \texttt{df[col].fillna(mean\_value,}} \colorbox{green!20}{\texttt{inplace=True)}.}
\item Return the modified \texttt{DataFrame}.
\end{enumerate}
\end{minipage}

  \end{tcolorbox}
  \caption{Side-by-side raw chains for BigCodeBench/889. Showing the verbatim steps clarifies how the +50\% step increase makes error handling, numeric-only selection, and the empty-CSV branch explicit, yielding a correct solution.}
  \label{fig:think-deeper-889-raw}
\end{figure}

%% file: fig/think_deeper_line.tex
\definecolor{Gemini25LightBlue}{HTML}{BFD8FF} \definecolor{QwenLightPurple}{HTML}{C9B6FF} \definecolor{ClaudeLightGreen}{HTML}{BFE7C6} \definecolor{O3MiniPink}{HTML}{F7B6C2} \definecolor{Gemini20Beige}{HTML}{E8D9B5} \definecolor{DeepseekGray}{HTML}{C7CBD6}

\begin{figure}[t!]
\centering
\begin{adjustbox}{max width=.98\linewidth}
\begin{tikzpicture}
\begin{groupplot}[
  group style={group size=1 by 2, vertical sep=26pt, ylabels at=edge left, xticklabels at=edge bottom},
  width=\linewidth,
  height=6.0cm,
  xmin=10, xmax=100,
  xtick={10,20,30,40,50,60,70,80,90,100},
  xlabel near ticks,
  ymin=0, ymax=0.40,
  ytick={0,0.1,0.2,0.3,0.4},
  ylabel={Resolution rate},
  tick label style={font=\small},
  grid=both,
  grid style={dashed,opacity=0.30},
  legend columns=3,
  legend cell align=left,
  legend style={/tikz/every even column/.append style={column sep=10pt}},
  every axis plot/.append style={
    line width=1.6pt,
    mark=*,
    mark size=2.3pt,
    mark options={solid,draw=black}
  }
]

\nextgroupplot[title={Full}, legend to name=leg:rr]

\addplot[color=ClaudeLightGreen, mark=*] coordinates {
(10,0.3442622951) (20,0.3934426230) (30,0.3442622951) (40,0.3606557377)
(50,0.3442622951) (60,0.3442622951) (70,0.3278688525) (80,0.3606557377)
(90,0.3442622951) (100,0.3442622951)
};
\addlegendentry{Claude-3.7-Sonnet-Thinking}

\addplot[color=DeepseekGray, mark=square*] coordinates {
(10,0.1851851852) (20,0.1666666667) (30,0.1851851852) (40,0.2037037037)
(50,0.1851851852) (60,0.1666666667) (70,0.1296296296) (80,0.2407407407)
(90,0.1481481481) (100,0.1851851852)
};
\addlegendentry{Deepseek-R1}

\addplot[color=Gemini20Beige, mark=triangle*] coordinates {
(10,0.22) (20,0.20) (30,0.18) (40,0.20)
(50,0.28) (60,0.28) (70,0.22) (80,0.24)
(90,0.28) (100,0.18)
};
\addlegendentry{Gemini-2.0-FT}

\addplot[color=Gemini25LightBlue, mark=star] coordinates {
(10,0.1636363636) (20,0.1636363636) (30,0.20) (40,0.0909090909)
(50,0.2545454545) (60,0.1818181818) (70,0.1090909091) (80,0.2363636364)
(90,0.1636363636) (100,0.2181818182)
};
\addlegendentry{Gemini-2.5-Flash}

\addplot[color=O3MiniPink, mark=diamond*] coordinates {
(10,0.0000000000) (20,0.0263157895) (30,0.0263157895) (40,0.0526315789)
(50,0.0263157895) (60,0.0263157895) (70,0.0263157895) (80,0.0526315789)
(90,0.0263157895) (100,0.0526315789)
};
\addlegendentry{O3-mini}

\addplot[color=QwenLightPurple, mark=o] coordinates {
(10,0.1176470588) (20,0.0392156863) (30,0.1176470588) (40,0.1372549020)
(50,0.0980392157) (60,0.1176470588) (70,0.0784313725) (80,0.0588235294)
(90,0.0980392157) (100,0.0980392157)
};
\addlegendentry{Qwen-QwQ}

\nextgroupplot[title={Hard}, xlabel={Percentage}]

\addplot[color=ClaudeLightGreen, mark=*] coordinates {
(10,0.0000000000) (20,0.1818181818) (30,0.2727272727) (40,0.0000000000)
(50,0.1818181818) (60,0.2727272727) (70,0.3636363636) (80,0.1818181818)
(90,0.1818181818) (100,0.1818181818)
};

\addplot[color=DeepseekGray, mark=square*] coordinates {
(10,0.0833333333) (20,0.1666666667) (30,0.1666666667) (40,0.0833333333)
(50,0.1666666667) (60,0.1666666667) (70,0.1666666667) (80,0.0833333333)
(90,0.0000000000) (100,0.1666666667)
};

\addplot[color=Gemini20Beige, mark=triangle*] coordinates {
(10,0) (20,0) (30,0) (40,0) (50,0) (60,0) (70,0) (80,0) (90,0) (100,0)
};

\addplot[color=Gemini25LightBlue, mark=star] coordinates {
(10,0.0833333333) (20,0.1666666667) (30,0.1666666667) (40,0.1666666667)
(50,0.0833333333) (60,0.1666666667) (70,0.1666666667) (80,0.0833333333)
(90,0.1666666667) (100,0.1666666667)
};

\addplot[color=O3MiniPink, mark=diamond*] coordinates {
(10,0.1111111111) (20,0.0000000000) (30,0.0000000000) (40,0.1111111111)
(50,0.1111111111) (60,0.1111111111) (70,0.0000000000) (80,0.2222222222)
(90,0.1111111111) (100,0.1111111111)
};

\addplot[color=QwenLightPurple, mark=o] coordinates {
(10,0.10) (20,0.20) (30,0.10) (40,0.20)
(50,0.20) (60,0.10) (70,0.10) (80,0.10)
(90,0.20) (100,0.00)
};

\end{groupplot}
\end{tikzpicture}
\end{adjustbox}

\pgfplotslegendfromname{leg:rr}

\caption{Resolution rate by model across step count percentage increase levels. Colors match the reference plot. Top: Full split. Bottom: Hard split.}
\label{fig:resolution_rate_full_hard}
\end{figure}
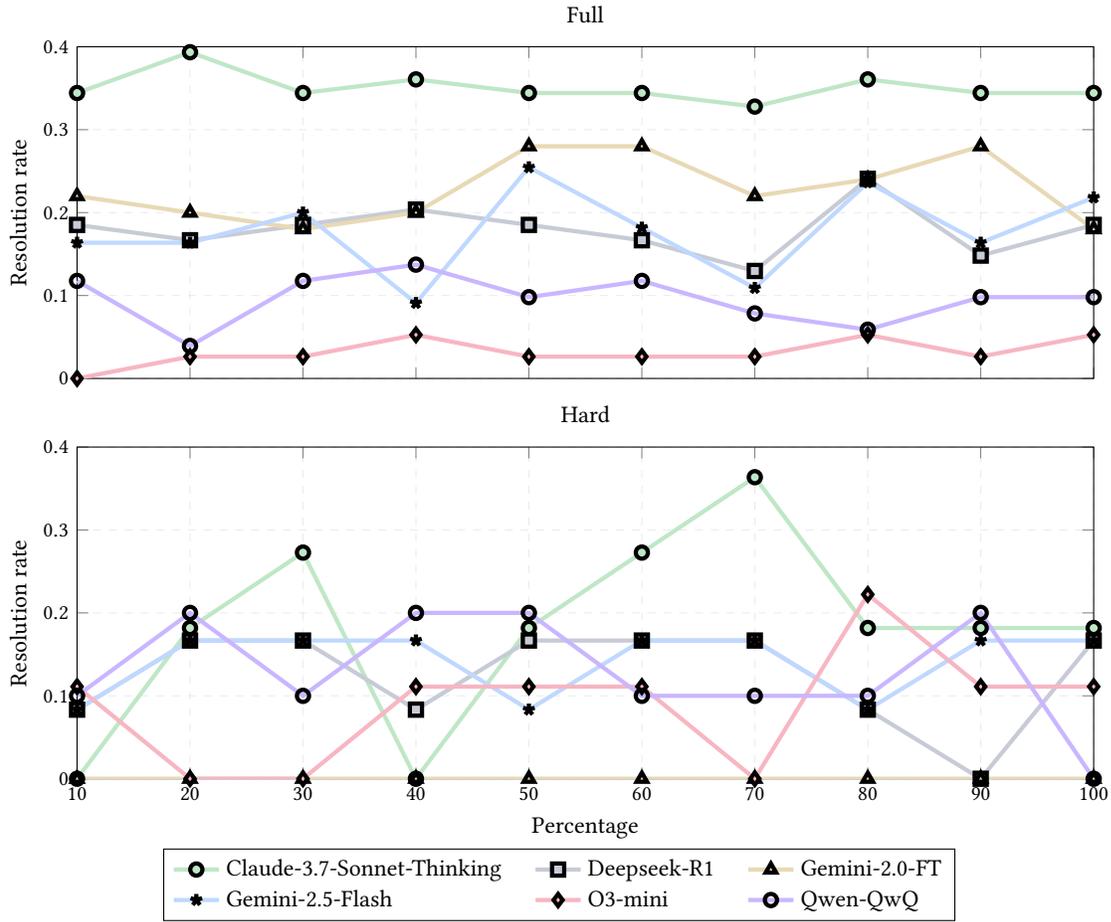

%% file: fig/1128-reduce-steps.tex
\begin{figure}[t]
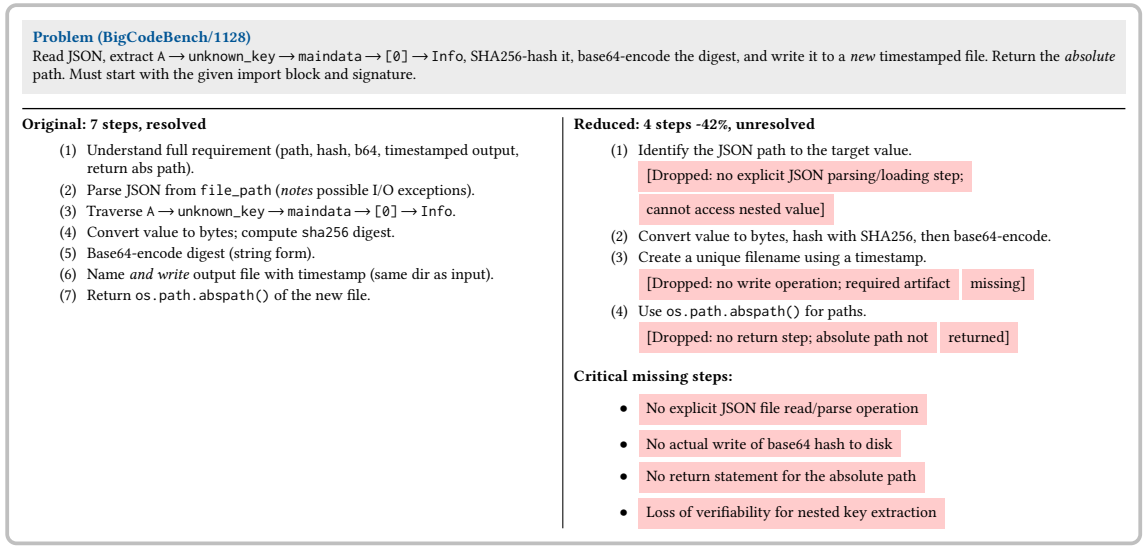

  \centering
  \begin{tcolorbox}[
      enhanced,
      width=\linewidth,
      boxsep=2pt,left=3pt,right=2.9pt,top=3pt,bottom=3pt,
      colback=white,colframe=black!25,
      fontupper=\scriptsize,
      before skip=2pt,after skip=2pt
  ]
  
\begin{tcolorbox}[
    colback=gray!15,
    colframe=gray!15,
    boxsep=2pt,
    left=2pt,right=2pt,top=2pt,bottom=2pt,
    sharp corners,
    boxrule=0pt
]
\href{https://huggingface.co/datasets/bigcode/bigcodebench-hard/viewer}{\textbf{Problem (BigCodeBench/1128)}}\quad
\\ 
Read JSON, extract \texttt{A\,$\rightarrow$\,unknown\_key\,$\rightarrow$\,maindata\,$\rightarrow$\,[0]\,$\rightarrow$\,Info}, SHA256-hash it, base64-encode the digest, and write it to a \emph{new} timestamped file. Return the \emph{absolute} path. Must start with the given import block and signature.
\end{tcolorbox}

\vspace{2pt}
\hrule height 0.5pt
\vspace{3pt}

\begin{minipage}[t]{0.48\linewidth}
\raggedright
\textbf{Original: 7 steps, resolved}\\[3pt]
\begin{enumerate}\setlength{\itemsep}{1pt}\setlength{\parskip}{0pt}\setlength{\topsep}{0pt}
\item Understand full requirement (path, hash, b64, timestamped output, return abs path).
\item Parse JSON from \texttt{file\_path} (\emph{notes} possible I/O exceptions).
\item Traverse \texttt{A\,$\rightarrow$\,unknown\_key\,$\rightarrow$\,maindata\,$\rightarrow$\,[0]\,$\rightarrow$\,Info}.
\item Convert value to bytes; compute \texttt{sha256} digest.
\item Base64-encode digest (string form).
\item Name \emph{and write} output file with timestamp (same dir as input).
\item Return \texttt{os.path.abspath()} of the new file.
\end{enumerate}
\end{minipage}%
\hfill\vline\hfill%
\begin{minipage}[t]{0.5\linewidth}
\raggedright
\textbf{Reduced: 4 steps -42\%, unresolved}\\[3pt]
\begin{enumerate}\setlength{\itemsep}{1pt}\setlength{\parskip}{0pt}\setlength{\topsep}{0pt}
\item Identify the JSON path to the target value. \colorbox{red!20}{[Dropped: no explicit JSON parsing/loading step;} \colorbox{red!20}{cannot access nested value]}
\item Convert value to bytes, hash with SHA256, then base64-encode.
\item Create a unique filename using a timestamp. \colorbox{red!20}{[Dropped: no write operation; required artifact} \colorbox{red!20}{missing]}
\item Use \texttt{os.path.abspath()} for paths. \colorbox{red!20}{[Dropped: no return step; absolute path not} \colorbox{red!20}{returned]}
\end{enumerate}
\vspace{3pt}
\textbf{Critical missing steps:}\\[2pt]
\begin{itemize}\setlength{\itemsep}{1pt}\setlength{\parskip}{0pt}\setlength{\topsep}{1pt}
\item \colorbox{red!20}{No explicit JSON file read/parse operation}
\item \colorbox{red!20}{No actual write of base64 hash to disk}
\item \colorbox{red!20}{No return statement for the absolute path}
\item \colorbox{red!20}{Loss of verifiability for nested key extraction}
\end{itemize}
\end{minipage}

  \end{tcolorbox}
  \caption{Side-by-side comparison for BigCodeBench/1128. The -42\% step reduction (7$\rightarrow$4) removes explicit file I/O operations (read and write) and the return step, turning key requirements into unmet assumptions and causing the solution to fail.}
  \label{fig:step-reduce-1128}
\end{figure}

%% file: fig/reduce_step.tex
\begin{figure}[t!]
\centering
\begin{adjustbox}{max width=.98\linewidth}
\begin{tikzpicture}
\begin{groupplot}[
  group style={group size=1 by 2, vertical sep=26pt, ylabels at=edge left, xticklabels at=edge bottom},
  width=\linewidth,
  height=5.8cm,
  xmin=0, xmax=90,
  xtick={0,10,20,30,40,50,60,70,80,90},
  ymin=0, ymax=1.02,
  ytick={0,0.2,0.4,0.6,0.8,1.0},
  tick label style={font=\small},
  grid=both,
  grid style={dashed,opacity=0.30},
  legend columns=3,
  legend cell align=left,
  legend style={/tikz/every even column/.append style={column sep=10pt}},
  every axis plot/.append style={
    const plot,
    line width=2.0pt,
    mark=*,
    mark size=2.5pt,
    mark options={solid,draw=black}
  }
]

\nextgroupplot[title={Full}, ylabel={Fraction still passing after cut $S(x)$}, legend to name=leg:survival]


\addplot[color=ClaudeLightGreen, mark=*] coordinates
{(0,1.000) (10,0.440) (20,0.160) (30,0.160) (40,0.160) (50,0.160) (60,0.120) (70,0.080) (80,0.040) (90,0.000)};
\addlegendentry{Claude-3.7-Sonnet-Thinking}

\addplot[color=DeepseekGray, mark=square*] coordinates
{(0,1.000) (10,0.250) (20,0.187) (30,0.156) (40,0.156) (50,0.156) (60,0.125) (70,0.094) (80,0.094) (90,0.000)};
\addlegendentry{Deepseek-R1}

\addplot[color=Gemini20Beige, mark=triangle*] coordinates
{(0,1.000) (10,0.584) (20,0.501) (30,0.307) (40,0.196) (50,0.140) (60,0.084) (70,0.084) (80,0.028) (90,0.000)};
\addlegendentry{Gemini-2.0-FT}

\addplot[color=Gemini25LightBlue, mark=star] coordinates
{(0,1.000) (10,0.838) (20,0.741) (30,0.483) (40,0.257) (50,0.096) (60,0.064) (70,0.064) (80,0.064) (90,0.032)};
\addlegendentry{Gemini-2.5-Flash}

\addplot[color=O3MiniPink, mark=diamond*] coordinates
{(0,1.000) (10,0.417) (20,0.250) (30,0.167) (40,0.167) (50,0.146) (60,0.125) (70,0.104) (80,0.083) (90,0.000)};
\addlegendentry{O3-mini}

\addplot[color=QwenLightPurple, mark=o] coordinates
{(0,1.000) (10,0.258) (20,0.229) (30,0.229) (40,0.229) (50,0.229) (60,0.172) (70,0.143) (80,0.086) (90,0.000)};
\addlegendentry{Qwen-QwQ}

\nextgroupplot[title={Hard}, xlabel={Relative step reduction (\%)}]

\addplot[color=ClaudeLightGreen, mark=*] coordinates
{(0,1.000) (10,0.000) (20,0.000) (30,0.000) (40,0.000) (50,0.000) (60,0.000) (70,0.000) (80,0.000) (90,0.000)};

\addplot[color=DeepseekGray, mark=square*] coordinates
{(0,1.000) (10,0.500) (20,0.500) (30,0.000) (40,0.000) (50,0.000) (60,0.000) (70,0.000) (80,0.000) (90,0.000)};

\addplot[color=Gemini20Beige, mark=triangle*] coordinates
{(0,1.000) (10,1.000) (20,1.000) (30,0.750) (40,0.750) (50,0.750) (60,0.500) (70,0.250) (80,0.000) (90,0.000)};

\addplot[color=Gemini25LightBlue, mark=star] coordinates
{(0,1.000) (10,0.000) (20,0.000) (30,0.000) (40,0.000) (50,0.000) (60,0.000) (70,0.000) (80,0.000) (90,0.000)};

\addplot[color=O3MiniPink, mark=diamond*] coordinates
{(0,1.000) (10,1.000) (20,0.600) (30,0.400) (40,0.400) (50,0.400) (60,0.400) (70,0.400) (80,0.200) (90,0.000)};

\addplot[color=QwenLightPurple, mark=o] coordinates
{(0,1.000) (10,0.000) (20,0.000) (30,0.000) (40,0.000) (50,0.000) (60,0.000) (70,0.000) (80,0.000) (90,0.000)};

\end{groupplot}
\end{tikzpicture}
\end{adjustbox}

\pgfplotslegendfromname{leg:survival}

\caption{Fraction of originally passed tasks that still pass after an $x\%$ reduction, with an added $0\%$ bin so each row sums to 1. Curves start at $S(0){=}1$ and decrease as $x$ grows.}
\label{fig:stepcut_survival}
\end{figure}
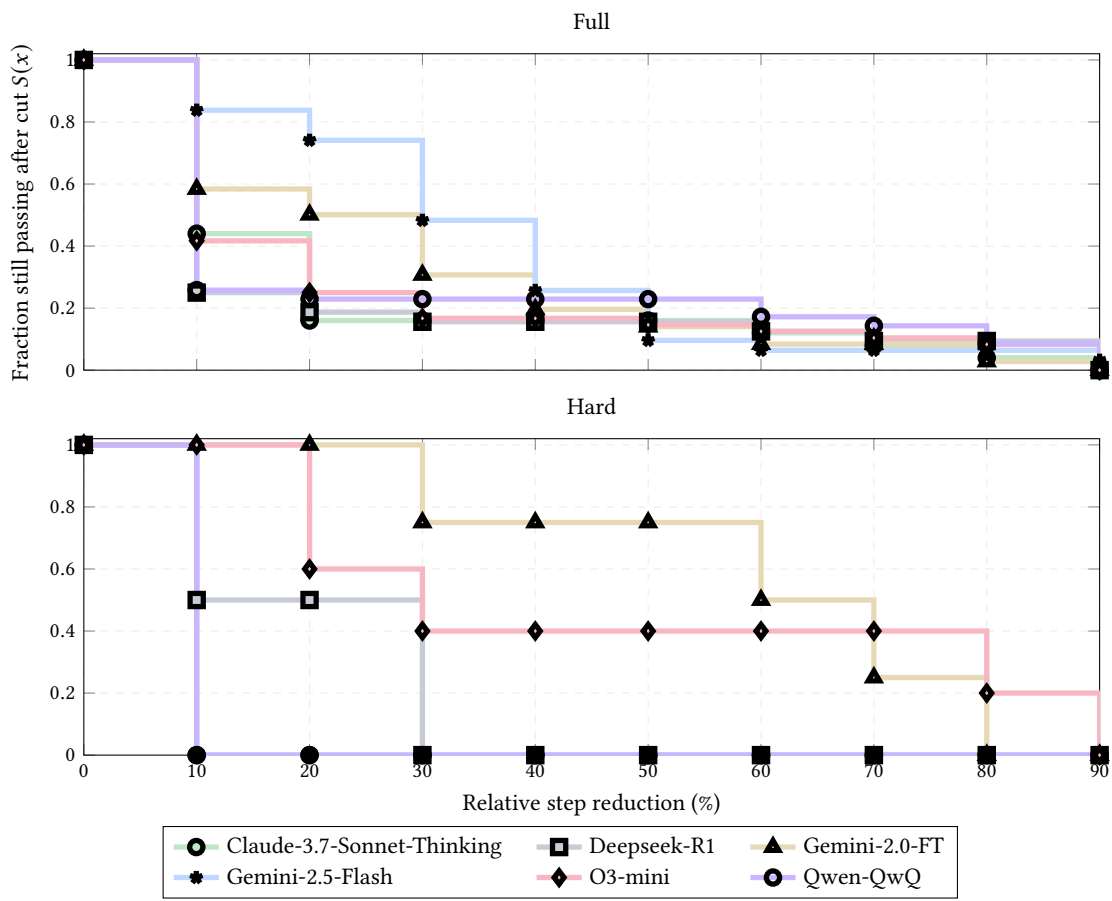

%% file: fig/verbosity_ana.tex
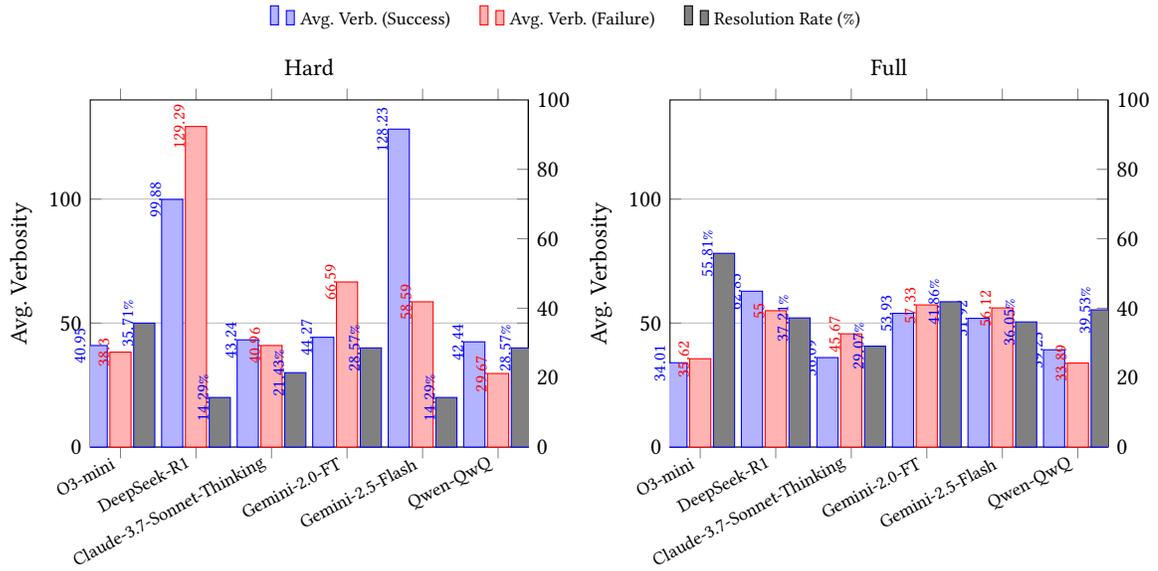
\begin{figure*}[t]
\centering
\pgfplotslegendfromname{verbLegend}
\vspace{.4em}

\begin{subfigure}{0.49\textwidth}
\centering
\begin{tikzpicture}
\begin{axis}[
  ybar,
  bar width=8pt,
  width=\textwidth, height=6.2cm,
  ymin=0, ymax=140,
  ytick={0,50,100},
  ylabel={Avg. Verbosity},
  title={Hard},
  symbolic x coords={O3-mini,DeepSeek-R1,Claude-3.7-Sonnet-Thinking,Gemini-2.0-FT,Gemini-2.5-Flash,Qwen-QwQ},
  xtick=data,
  xticklabel style={rotate=28,anchor=east, font=\footnotesize},
  enlarge x limits=0.08,
  ymajorgrids,
  legend to name=verbLegend,
  legend columns=3,
  legend style={/tikz/every even column/.append style={column sep=1em}, draw=none, font=\footnotesize},
]
  \addplot+[ybar, bar shift=-9pt,
            nodes near coords,
            every node near coord/.append style={
              font=\scriptsize, rotate=90, anchor=south, yshift=1pt
            }] coordinates
    {(O3-mini,40.95) (DeepSeek-R1,99.875) (Claude-3.7-Sonnet-Thinking,43.24)
     (Gemini-2.0-FT,44.27) (Gemini-2.5-Flash,128.225) (Qwen-QwQ,42.44)};
  \addlegendentry{Avg. Verb. (Success)}

  \addplot+[ybar, bar shift=0pt,
            nodes near coords,
            every node near coord/.append style={
              font=\scriptsize, rotate=90, anchor=south, yshift=1pt
            }] coordinates
    {(O3-mini,38.30) (DeepSeek-R1,129.29) (Claude-3.7-Sonnet-Thinking,40.96)
     (Gemini-2.0-FT,66.59) (Gemini-2.5-Flash,58.59) (Qwen-QwQ,29.67)};
  \addlegendentry{Avg. Verb. (Failure)}

  \addlegendimage{ybar, fill=gray}
  \addlegendentry{Resolution Rate (\%)}
\end{axis}

\begin{axis}[
  ybar,
  bar width=8pt,
  width=\textwidth, height=6.2cm,
  ymin=0, ymax=100,
  axis y line*=right,
  axis x line=none,
  symbolic x coords={O3-mini,DeepSeek-R1,Claude-3.7-Sonnet-Thinking,Gemini-2.0-FT,Gemini-2.5-Flash,Qwen-QwQ},
  xtick=data,
  enlarge x limits=0.08,
]
  \addplot+[ybar, bar shift=+9pt, fill=gray,
            nodes near coords,
            every node near coord/.append style={
              font=\scriptsize, rotate=90, anchor=south, yshift=1pt
            },
            point meta=y,
            nodes near coords={\pgfmathprintnumber[fixed,precision=2]{\pgfplotspointmeta}\%}
           ] coordinates
    {(O3-mini,35.71) (DeepSeek-R1,14.29) (Claude-3.7-Sonnet-Thinking,21.43)
     (Gemini-2.0-FT,28.57) (Gemini-2.5-Flash,14.29) (Qwen-QwQ,28.57)};
\end{axis}
\end{tikzpicture}
\end{subfigure}
\hfill
\begin{subfigure}{0.49\textwidth}
\centering
\begin{tikzpicture}
\begin{axis}[
  ybar,
  bar width=8pt,
  width=\textwidth, height=6.2cm,
  ymin=0, ymax=140,
  ytick={0,50,100},
  ylabel={Avg. Verbosity},
  title={Full},
  symbolic x coords={O3-mini,DeepSeek-R1,Claude-3.7-Sonnet-Thinking,Gemini-2.0-FT,Gemini-2.5-Flash,Qwen-QwQ},
  xtick=data,
  xticklabel style={rotate=28,anchor=east, font=\footnotesize},
  enlarge x limits=0.08,
  ymajorgrids,
  legend=false, 
]
  \addplot+[ybar, bar shift=-9pt,
            nodes near coords,
            every node near coord/.append style={
              font=\scriptsize, rotate=90, anchor=south, yshift=1pt
            }] coordinates
    {(O3-mini,34.01) (DeepSeek-R1,62.85) (Claude-3.7-Sonnet-Thinking,36.09)
     (Gemini-2.0-FT,53.93) (Gemini-2.5-Flash,51.92) (Qwen-QwQ,39.23)};

  \addplot+[ybar, bar shift=0pt,
            nodes near coords,
            every node near coord/.append style={
              font=\scriptsize, rotate=90, anchor=south, yshift=1pt
            }] coordinates
    {(O3-mini,35.62) (DeepSeek-R1,55.00) (Claude-3.7-Sonnet-Thinking,45.67)
     (Gemini-2.0-FT,57.33) (Gemini-2.5-Flash,56.12) (Qwen-QwQ,33.89)};
\end{axis}

\begin{axis}[
  ybar,
  bar width=8pt,
  width=\textwidth, height=6.2cm,
  ymin=0, ymax=100,
  axis y line*=right,
  axis x line=none,
  symbolic x coords={O3-mini,DeepSeek-R1,Claude-3.7-Sonnet-Thinking,Gemini-2.0-FT,Gemini-2.5-Flash,Qwen-QwQ},
  xtick=data,
  enlarge x limits=0.08,
]
  \addplot+[ybar, bar shift=+9pt, fill=gray,
            nodes near coords,
            every node near coord/.append style={
              font=\scriptsize, rotate=90, anchor=south, yshift=1pt
            },
            point meta=y,
            nodes near coords={\pgfmathprintnumber[fixed,precision=2]{\pgfplotspointmeta}\%}
           ] coordinates
    {(O3-mini,55.81) (DeepSeek-R1,37.21) (Claude-3.7-Sonnet-Thinking,29.07)
     (Gemini-2.0-FT,41.86) (Gemini-2.5-Flash,36.05) (Qwen-QwQ,39.53)};
\end{axis}
\end{tikzpicture}
\end{subfigure}

\caption{Average reasoning verbosity (success/failure) and resolution rate for each model, split by difficulty. Left y-axes share the same scale and tick interval (0, 50, 100). Resolution rate uses the right y-axis.}
\label{fig:verbosity-resolution-bars}
\end{figure*}

%% file: sec/resultsanalysis.tex
\section{Assessing the Reasoning Quality in Thinking Code LLMs (RQ2)}

While recent work has evaluated the reasoning content produced by thinking LLMs~\cite{qiu2025quantifyingreasoningabilitiesllms}, it does not establish a structured classification of reasoning failures that could provide a qualitative, cross-model taxonomy of error patterns grounded in human evaluation. Consequently, important questions remain unanswered, for example, what are the common patterns of problematic reasoning? 
This gap is particularly significant in software engineering contexts, where reasoning quality directly impacts code correctness and reliability. To address this gap, we build a comprehensive taxonomy to present the common patterns that occur in problematic reasoning. In addition, building on insights from prior work, step-by-step reasoning can be characterized along three core dimensions: Efficiency~\cite{xia2025evaluating}, Logical Consistency~\cite{lee2025evaluatingstepbystepreasoningtraces}, and Completeness~\cite{brassard2024acorn}. Together, these dimensions provide a literature-backed, sufficient basis for reasoning quality, asking whether the chain is concise, sound, and thorough. 
In this work, we employ these three criteria to evaluate the quality of the reasoning process. For each criterion, we define a metric as follows: efficiency (\ref{rq2.2}), logic consistency (\ref{rq2.3}), and completeness (\ref{rq2.4}). Fig. \ref{fig:flowchart} illustrates our workflow for the annotation work. We approach reasoning quality through several analyses: first, establishing a comprehensive taxonomy of failure patterns (\ref{rq2.1}), then systematically evaluating reasoning quality across three dimensions: efficiency (\ref{rq2.2}), logic consistency (\ref{rq2.3}), and completeness (\ref{rq2.4}). And we formalize our research questions as follows:

\input{fig/flowchart-humaneval}

\begin{itemize}
    \item \textbf{RQ2.1:} What patterns characterize problematic (faulty) reasoning?
    \item \textbf{RQ2.2:} How efficient are thinking LLMs' reasoning traces? 
    \item \textbf{RQ2.3:} How logically consistent are the reasoning traces? 
    \item \textbf{RQ2.4:} How complete are the reasoning traces? 
    \item \textbf{RQ2.5:} How does task complexity (Hard vs.\ Full) affect (i) reasoning quality (efficiency, logic, and completeness) and (ii) solution correctness across models?
\end{itemize}

\subsection{Human Evaluation Setup and Methodology} 
To systematically assess reasoning quality from developers' perspectives, we conducted a comprehensive human evaluation study with 21 participants. This section describes our rigorous evaluation protocol, which includes participant recruitment, the development of standardized coding guidelines, training procedures, and an inter-rater reliability analysis. Figure \ref{fig:humanwork_process} illustrates the overall workflow of our participant recruitment and evaluation process. The process consists of four main phases: (1) Participant Recruitment, (2) Training Session, (3) Task Assignment, and (4) Evaluation Phase. We describe each process in detail below. 
\input{fig/humanwork}

\subsubsection{Participants Recruitment}
To conduct a systematic assessment, we recruited 21  graduate students with backgrounds in computer science, all of whom were familiar with Python programming and the use of Large Language Models (LLMs). 
\subsubsection{Development of the Coding Guide}
Before participant training, we developed a coding guide to standardize the evaluation process. The coding guide was created through multiple rounds of discussion and refinement among three researchers. We established clear evaluation criteria, systematic procedures, and provided concrete examples to guide participants. 
The coding guide consists of three main evaluation criteria, each with detailed evaluation procedures. 

\begin{itemize}
    \item \textbf{Efficiency:} This metric measures whether each reasoning step contributed meaningfully toward solving the problem, identifying redundant steps that don't advance the solution. The evaluation procedure requires participants to count the total number of reasoning steps they take. And for each reasoning step, determine if it is: 
        \begin{itemize}
            \item \textbf{Essential:} Steps that introduce necessary information, perform required calculations, or make crucial connections needed for solving the problem.
            \item \textbf{Non-Essential:} Steps that are redundant or provide unnecessary elaboration that do not advance the solution.
        \end{itemize}
        
    \item \textbf{Logic Consistency:} This metric assesses the coherence and internal consistency of the reasoning process. For each step (except the first), participants determine if it logically follows from previous steps. Steps are marked as ``coherent'' if the logical connection is clear and valid, or ``incoherent'' if there are logical gaps and inconsistencies.

    \item \textbf{Completeness:} This metric checks whether the reasoning thoroughly addresses all aspects of the problem. This procedure requires participants to carefully read the problem description and identify all essential requirements stated therein. They then check if each component is explicitly addressed in the reasoning, marking each as either 'addressed' or 'not addressed'. 
    
    Participants are also expected to verify whether edge cases and constraints were properly considered in the reasoning and list any missed edge cases in handling.
\end{itemize}

\subsubsection{Training Session}
Before the evaluation, participants underwent a training session (approximately 10 minutes) during which we introduced the study objectives, explained the three evaluation criteria, and provided annotated examples of reasoning traces. This training was intended to calibrate evaluators and ensure consistent application of the assessment guidelines. 

\subsubsection{Task Assignment}
Each participant was then assigned a specific subset of 10 tasks, sampled across six reasoning LLMs. The sampling strategy ensured that each task was assigned to exactly three different participants, enabling inter-rater reliability analysis. 

\subsubsection{Evaluation Phase}
The evaluation phase constituted the main data collection period, during which participants independently assessed the assigned reasoning contents using the coding guide. On average, participants spent approximately 12–15 minutes per task, resulting in a total time commitment of about 2–2.5 hours per participant. This procedure ensured consistent assessment standards while enabling a comprehensive analysis of reasoning quality patterns across different models and task complexities.

\subsubsection{Inter-Rater Agreement Analysis}
\input{tab/icc_scale}
\input{tab/overallicc}
To assess the reliability of our evaluation methodology and the consistency of participant annotations, we computed inter-rater agreement using Intraclass Correlation Coefficient (ICC) \cite{shrout1979intraclass}. For each reasoning content and each criterion, we derived numeric quality scores from the evaluation.

We used the $ICC(1,k)$ model, which is appropriate for our design, where each task is rated by $k=3$ different participants randomly selected. Specifically, for a one-way random-effects ANOVA with $n$ tasks and $k$ raters per task, the average-measures ICC is
\[
\mathrm{ICC}(1,k)=\frac{MS_{B}-MS_{W}}{MS_{B}},
\]
where $MS_{B}$ is the one-way ANOVA mean square between tasks (targets/rows) and $MS_{W}$ is the within-task (residual/error) mean square computed from the same ANOVA; $k$ denotes the number of raters per task (here $k=3$) and $n$ the number of tasks. We computed separate ICC scores for each of the three evaluation criteria. We applied the standard interpretation scale for ICC values, which is shown in Table \ref{tab:icc-bands}. Values above 0.9 indicate excellent agreement, 0.75--0.9 indicate good agreement, 0.5--0.75 indicate moderate agreement, and below 0.5 indicate poor agreement \cite{koo2016guideline}. Our analysis demonstrates strong inter-rater reliability across all three evaluation criteria: Efficiency (ICC = 0.926), Logic Consistency (ICC = 0.934), and Completeness (ICC = 0.863), as shown in Table \ref{tab:icc-score}. 

Having established our evaluation methodology, we now present our analysis of reasoning quality across the three dimensions, beginning with a taxonomy of problematic reasoning patterns.

\subsection{RQ2.1 Patterns in Problematic (faulty) Reasoning}
\label{rq2.1}

\input{fig/tax}

\input{fig/737-incomplete}

\input{fig/1082-example}
\input{fig/458-example}
\input{fig/336-example-specific}
\input{fig/859-example-enforced}
\subsubsection{Approach}
Building on the reasoning process analysis in RQ1, which involves step and verbosity count, we aim to further investigate the quality of reasoning in this section. We study the same six thinking LLMs (i.e., DeepSeek-R1, OpenAI-o3-mini, Claude 3.7 Sonnet-Thinking, Gemini-2.0-Flash-Thinking, Gemini-2.5-Flash, and Qwen-QwQ) on the 100 BigCodeBench-Instruct tasks, and 21 trained annotators assess each chain along three criteria—Efficiency, Logic Consistency, and Completeness. Our empirical study begins by examining the common patterns of faulty reasoning. Through a detailed analysis of reasoning failures across six evaluated thinking LLMs, we derive a taxonomy that captures the fundamental patterns of reasoning deficiencies in code generation tasks (as shown in Figure~\ref{fig:taxonomy_of_issue_patterns}). 
The taxonomy development followed a rigorous qualitative methodology grounded in open-coding techniques. Initially, all authors independently examined a subset of failure instances identified by our 21 human evaluators across the three metrics (reasoning efficiency, reasoning logic consistency, and reasoning completeness). Through iterative open-coding sessions, we systematically identified recurring patterns and failure modes in the reasoning processes. The authors then collaborated in multiple rounds of discussion and refinement to consolidate similar patterns, resolve disagreements, and establish clear definitional boundaries for each category. This collaborative coding process involved constant comparison analysis, where newly identified patterns were continuously compared against existing categories to ensure theoretical saturation and category distinctiveness. The final taxonomy represents a consensus-driven classification system that emerged from this systematic qualitative analysis, providing a structured framework for understanding how thinking LLMs systematically fail in their reasoning processes across different dimensions of reasoning quality.

\subsubsection{Results}
Guided by the three annotation metrics: efficiency, logic consistency, and completeness, we organize the observed reasoning problematic patterns into three top-level categories: \textbf{Completeness Issue}, \textbf{Efficiency Issue}, and \textbf{Logic Consistency Issue}. For each component, we define second-level subcategories that capture recurring failure patterns. Where warranted, we further refine these into third-level sub-subcategories that describe concrete manifestations observed in the traces. We introduce each pattern below, with precise definitions and representative examples to anchor the taxonomy.

\noindent \textbf{Completeness Issues:} represent the most critical and frequent reasoning deficiencies (44.5\%), encompassing systematic gaps in problem analysis and reasoning coverage. 

\begin{enumerate}
    \item \textbf{Cut-off Reasoning:} manifests when models prematurely terminate their reasoning chains, often after identifying surface-level requirements and during the middle of the reasoning chain, but fail to pursue the whole logical progression necessary for comprehensive problem-solving. This pattern typically occurs when models recognize the general problem domain but lack the persistence or capability to work through all necessary reasoning steps, resulting in partial solutions that appear plausible but miss essential components. The example is presented in Fig. \ref{fig:incomplete-737}. 
    
    \item \textbf{Missing Coverage of Requirements:} represents a more fundamental failure where models overlook explicitly stated problem constraints, functional requirements, or specific behaviors. It includes three sub-categories, i.e., a) \textbf{Problem Constraints Not Enforced}, where the reasoning fails to enforce prompt-stated input assumptions, schema checks, parameter exactness, scope constraints, or required runtime/reproducibility settings that gate correct execution. Fig. \ref{fig:missing-constraints-859} shows an example of this category; (b) \textbf{Functional Requirement Omitted}, where a required transformation, algorithm, or computation is not performed. Fig. \ref{fig:missing-coverage-458} shows an example of this category; and (c) \textbf{Miss-covering Specified Behavior}, where the produced artifacts not follow the prompt’s interface or presentation specification, such as returns, files, names, labels, or write options, Fig. \ref{fig:missing-covering-336} shows an example for this category.  
    
    \item \textbf{Lack of Edge Case Handling:} emerges as the most prevalent and technically significant failure pattern, with four distinct subcategories that reveal weaknesses in robust code generation reasoning. It has four sub-categories, i.e., a) \textbf{File I/O} encompasses failures to consider file access permissions, non-existent file paths, corrupted file formats, and concurrent file access scenarios that are essential for real-world code reliability. b) \textbf{Value Validation} includes insufficient consideration of input sanitation, type checking, format validation, and range verification that are crucial for preventing runtime errors and security vulnerabilities. c) \textbf{Boundary Condition} includes handling of $\min/\max$ values, $\{\varnothing, 0, \texttt{None}\}$ value processing, special numeric values like $\{\texttt{NaN}, \pm\infty\}$ management that frequently cause production code failures (see Fig. \ref{fig:lacking-edges-1082} as an example). d) \textbf{Resource Limits} addresses scalability concerns, including large memory allocation scenarios, extensive directory traversal operations that are essential for production-ready code generation. 
\end{enumerate}

\input{fig/1006-example-structural}
\input{fig/453-restatement}

\input{fig/797-example-circular}

\input{fig/473-example}

\input{fig/805-example-gap}
\input{fig/1036-example-claim}
\input{fig/906-example-execution}
\input{fig/1029-example-data}
\input{fig/586-example-seq}

\noindent \textbf{Efficiency Issues:} captures reasoning processes that, while potentially arriving at correct conclusions, demonstrate systematic inefficiencies that undermine practical reasoning quality. 
\begin{enumerate}
    \item \textbf{Redundancy:} manifests as unnecessarily lengthy reasoning chains that include repetitive explanations, redundant problem restatements, and excessive elaboration that obscures rather than clarifies the reasoning process. We defined two sub-subcategories under this: a) \textbf{Redundant Structural Walkthrough}, which represents a step that re-describes the plan, structure, or approach the model already established in the previous steps. These steps merely recapitulate existing reasoning organization without advancing the solution. Fig. \ref{fig:redundancy-1006} shows an example of this category; b) \textbf{Restatement Provided Information} occurs when the model repeats information already explicitly stated in the problem description or task specification. Fig. \ref{fig:redundancy-453} shows an example of this category. 

\input{fig/distribution}

    \item \textbf{Overthinking:} 
    captures instances where models engage in excessive, unproductive analysis that fails to advance problem-solving progress. This manifests in two distinct patterns: circular deliberation without resolution and expansive reasoning beyond the problem scope. We identify two subcategories under this: a) \textbf{Self-debate}, which occurs when the model becomes trapped in recursive loops, repeatedly revisiting the same set of alternatives without adding new criteria or making a decision. These circular deliberations consume computational resources and reasoning steps, yet fail to reach conclusions or advance the solution, representing cognitive spinning that produces no forward progress. The example is presented in Fig. \ref{fig:circular-797}; b) \textbf{Beyond requirement}. represents where the models introduce considerations, constraints, or requirements that are not present in the original problem specification. This pattern often manifests as models fabricating additional complexity, assuming unstated requirements, or addressing problems that extend beyond the specified scope. The example is provided in Fig. \ref{fig:beyond-reqs-473}. 
\end{enumerate}

\noindent \textbf{Logic Issues:}  represent the most fundamentally problematic reasoning deficiencies, as they undermine the basic reliability of the reasoning process regardless of other quality factors. 
\begin{enumerate}
    \item \textbf{Logic Inconsistency:} encompasses internal contradictions within reasoning chains, where models make statement or assumptions that conflicts with previous reasoning steps, maintain incompatible definitions or approaches simultaneously, or draw conclusions that logically contradict their own premises. While relatively infrequent in our evaluation, these failures represent catastrophic reasoning breakdowns that completely undermine the reliability of the solution. We identify three subcategories under this: a) \textbf{Logic Gap} occurs when reasoning steps contain unjustified leaps, missing intermediate inferences, or unsupported assumptions where concepts are introduced or advanced without sufficient logical foundation from the prior step. The model may reference ideas, constraints, or approaches that lack adequate explanation or justification, creating discontinuities in the reasoning chain that undermine its validity. Fig. \ref{fig:logic-gap-805} shows an example for this category; b) \textbf{Claim-Implementation Mismatch} manifests when a model explicitly claims an approach or plan in its reasoning but subsequently implements something different or contradictory in later steps. This creates an internal inconsistency where the articulated reasoning strategy diverges from the actual solution patch followed. Fig. \ref{fig:logic-inconsistency-1036} shows an example of this category; c) \textbf{Sequence Misorder} represents violations of necessary logical or operational ordering, where the model proposes or reasons about steps in an order that contradicts dependencies, prerequisites, or temporal constraints inherent to the problem. Fig. \ref{fig:logic-misorder-586} shows an example for this category. 
    
    \item \textbf{Requirement Misinterpretation:} captures the misunderstanding or incorrect parsing of problem statements. We identify three subcategories under this: a) \textbf{Execution \& Behavior Misread}: when the expected runtime behavior or procedural setup is misunderstood. Fig. \ref{fig:req-misread-906} shows an example of this category; b) \textbf{Data \& Constraints Misread}: when the data semantics or constraints are misread. Fig. \ref{fig:req-misread-1029} provides an example of this category.
\end{enumerate}

We present the distribution at each subcategory level, which shows the reasoning deficiency profiles across the six thinking LLMs as shown in Fig. \ref{fig:category-and-subsub-combined}. Furthermore, to present more specific and detailed problematic patterns and their distribution, we also present the distribution at each sub-subcategory level, as shown in Fig. \ref{fig:category-and-subsub-combined}. At the subcategory level shown in Fig. \ref{fig:category-and-subsub-combined}, the lack of edge cases handling dominates at 32.17\% of all problematic patterns. Under this, missing value validation is the major concern with 53.89\% proportion among all edge case conditions (38.95\% among all completeness issues), followed by boundary condition (22.28\%) (16.10\% among all completeness issues), as shown in Fig. \ref{fig:category-and-subsub-combined}. A second major problematic pattern is from efficiency, where redundancy accounts for 23.33\% of all problematic cases. Among these, 30.71\% comes from Gemini-2.5-Flash, followed by Gemini-2.0-FT, with a proportion of 27.86\% under the redundancy category. 

Other interesting observations are that DeepSeek-R1 more often encounters the overthinking problem, specifically, it is more likely to exhibit a self-debate problem than other models (66.67\%), indicating that it is more often trapped and lost in its own reasoning chain, with an 8.33\% frequency of all its problematic reasoning contents. Gemini-2.5-Flash is more likely to produce cut-off reasoning, where the reasoning chain is cut off in the middle of the sentence, with a 66.67\% proportion among all cut-off reasoning cases we collected, and this accounts for 26.87\% of all problematic reasoning in Gemini-2.5-Flash. 

\begin{observation}{}
{}Completeness issues are the dominant deficiency (44.5\%), especially missing edge cases. Efficiency-relevant problems are secondary (33.5\%). More interestingly, DeepSeek-R1 exhibits overthinking and self-debate (8.33\% of its failures), while Gemini-2.5-Flash frequently produces incomplete cut-off reasoning (26.87\% of its failures).
\end{observation}

\subsection{Metric Evaluation}

Having identified the patterns of problematic reasoning through our taxonomy in Section~\ref{rq2.1},  we now systematically quantify reasoning quality across three categories: efficiency, logic consistency, and completeness. To enable integrated assessment, we introduce R\_3dim as an aggregate quality metric, computed as the arithmetic mean of the three individual dimension scores for each reasoning trace. This composite metric provides an overall quality indicator while preserving the individual dimension scores for detailed analysis and evaluation. Table \ref{tab:overall_quality} summarizes the reasoning quality across all models and tasks, presenting mean scores and standard deviations for each dimension as well as the overall R\_3dim metric. We discuss the analysis for each criterion in the following sections.

\subsubsection{RQ2.2 Efficiency Evaluation}
\label{rq2.2}
\input{fig/eff_example}
\paragraph{Approach}

For the efficiency criterion, we evaluate whether each reasoning step contributes meaningfully toward problem resolution rather than providing redundant elaboration or repetition. Annotators label steps Essential, which introduce necessary information, perform required calculations, establish crucial logical connections, or advance progress, or Non-essential, which repeat prior content, add unnecessary detail, or do not advance the solution. To make the annotation protocol concrete, Figure~\ref{fig:eff_example} shows a worked example where annotators evaluate a reasoning chain against the task specification. In this example, annotators first align on the problem description (left), then review each step of the reasoning process and label it using the Efficiency criteria (right). Step 1 addresses requirements comprehension, step 2 covers approach design, and step 3 provides implementation details that directly realize the requirements. These steps are marked Essential because each introduces necessary information that advances the solution toward the required tuple and plot. Step 4 gives a verification checklist that is marked as Non-essential, as there is no contribution information advanced to solve the problem; therefore, it is marked as Non-Essential.  

We define the Efficiency Score metric as follows:
\[
\mathrm{\textbf{Efficiency Score}}=\frac{\mathrm{\text{Number of Essential Steps}}}{\text{Total Number of Steps}}
\]
Applying our metric calculation, the efficiency score for this example is $\tfrac{3}{4}=0.75$.
\paragraph{Results}
Across 600 tasks (100 tasks $*$ 6 models), efficiency issues appear in 33.50\% of cases, indicating that while many chains are concise, non-progressing steps remain common. Table. \ref{tab:overall_quality} presents efficiency scores for all six models. High efficiency scores here indicate that the most steps make measurable progress on the task. The results reveal remarkably high efficiency across most models, with mean scores ranging from 0.796 to 0.975. Three models:  OpenAI-o3-mini $(0.971 \pm 0.049)$, Gemini-2.0-Flash-Thinking $(0.949 \pm 0.077)$, and DeepSeek-R1 $(0.948 \pm 0.131)$, exhibit excellent efficiency with low variance, indicating consistent, focused reasoning across tasks. In contrast, Gemini-2.5 $(0.796 \pm 0.363)$ shows substantially lower efficiency. This is consistent with our observations of problematic patterns observed in \ref{rq2.1}, which include the frequent cut-off reasoning and redundant elaboration in its reasoning contents. These patterns also align with our previous findings: when step counts are adjusted, step reduction generally lowers resolution, especially on \textit{Hard} tasks, and only small cuts are safe when the original chain already covers the stated requirements. By contrast, increasing steps did not uniformly help the models. Overall, the data support that efficiency, under our human labels, captures how much of the chain materially advances the solution, not merely how long the chain is. 

\begin{observation}{}
{}Efficiency issues affect 33.5\% of reasoning chains, with Gemini-2.5-Flash accounting for 25.87\% of all efficiency problems. This also align with our evaluation results that Gemini-2.5-flash has the lowest human evaluation efficiency score $(0.796 \pm 0.363)$.
\end{observation}

\subsubsection{RQ2.3 Logic Consistency Evaluation}
\label{rq2.3}

\paragraph{Approach}
We assess whether each step follows from previous steps without gaps or invalid inferences. Annotators judge consecutive transitions within the chain as coherent or not. The coherent score is:

\[
\mathrm{\textbf{Coherence Score}}=\frac{\mathrm{\text{Number of Coherent Steps}}}{\text{Total Number of Steps - 1}}
\]

To make the annotation protocol concrete, Figure~\ref{fig:logic_example} illustrates a labeling example in which annotators evaluate a reasoning chain. In this example, step 1 correctly states the task and requirements; step 2 commits to an unconditional plan that converts the input list to a numpy array, computes the FFT, and then plots the diagram. However, step 3 then introduces an early return for empty input (return before FFT) and prepares magnitudes otherwise, which conflicts with step 2; therefore, step 3 is labeled incoherent. The other steps are coherent, from step 1 to step 2, which transitions from requirements to implementation design, and the transition from step 3 to step 4 involves magnitude preparation, plotting, and returning outputs. Applying our metric gives a coherence score of $\tfrac{2}{3}=0.67$. 
\input{fig/logic_example}

\paragraph{Results}
Logic Consistency issues occurs in 7.5\% of all tasks, indicating that most chains maintain step-by-step coherence. Against this backdrop, Table \ref{tab:overall_quality} reveals strong, consistent performance across most models. o3-mini ($0.976\pm0.038$), Gemini-2.0-FT ($0.969\pm0.059$), Claude-3.7-Sonnet-Thinking ($0.963\pm0.066$), and DeepSeek-R1 ($0.965\pm0.096$) are closely clustered at the top, with few incoherent transitions and relatively small variability. Qwen-QwQ ($0.930\pm0.206$) is lower with a greater spread, and Gemini-2.5 ($0.842\pm0.316$) again lags and varies widely across tasks. We observed that outright logical breaks are uncommon for most models and, therefore, not the primary source of variability in final correctness on tasks.

\begin{observation}{}
{}Logic consistency is generally strong across the thinking LLMs, with issues occurring in only 7.5\% of tasks. Most models maintain high coherence scores (>0.96), indicating that step-to-step logical flow is not a primary bottleneck. However, the high variance in certain models (e.g., Gemini-2.5: $0.842 \pm 0.316$) suggests that when logical breaks do occur, they tend to be model-specific weaknesses rather than inherent challenges of the reasoning tasks themselves.
\end{observation}

\subsubsection{RQ2.4 Completeness Evaluation}
\label{rq2.4}
\paragraph{Approach}
We measure whether the chain addresses all stated requirements and verifies the necessary conditions. Annotators enumerate the requirements set from the prompt and mark each item as addressed or missed, as shown in a concrete annotation work example presented in Figure \ref{fig:completeness_example}. The annotators first enumerate four core explicit requirements from the problem description: (1) generate a random string of length within the given bounds; (2) use characters from 'letters'; (3) evaluate the similarity; (4) return the output of a tuple. The annotators then scan the reasoning chain and mark a requirement as addressed when the chain explicitly commits to the action. In the given example, the model's reasoning steps 1-3 address the requirements (1) to (3) by stating the bounded length, the 'letters' usage, and the similarity computation. However, the chain never states a tuple return; therefore, requirement (4) is marked as non-addressed. We computed the completeness score as:

\[
\mathrm{\textbf{Completeness Score}}=\frac{\mathrm{\text{Requirements Addressed}}}{\text{Total Requirements}}
\]
In this case, the resulting completeness score is $\tfrac{4}{5}=0.8$. 
\input{fig/completeness_example}

\paragraph{Results}
Completeness issues are prevalent in our corpus, with 44.5\% of tasks involving completeness problems, making them the dominant source of failure. 
This prevalence informs the interpretation of the model-level completeness scores that follow. Gemini-2.0-Flash-Thinking achieves exceptional performance at $0.994 \pm 0.024$, representing the highest score across all models and metrics, with remarkably low variance, suggesting robust and comprehensive reasoning across diverse task types. Followed by DeepSeek-R1 ($0.989\pm0.041$), o3-mini ($0.977\pm0.033$), and Claude-3.7-Sonnet-Thinking ($0.958\pm0.058$).  Gemini-2.5 exhibits the weakest completeness, with a score of $0.830 \pm 0.339$, which is the lowest across all metrics in our evaluation.

The errors underlying lower Completeness Scores are consistent with our qualitative review: missed requirements and insufficient treatment of stated conditions or edge cases, rather than purely local contradictions. This also explains the interaction with our step-adjustment experiments: where completeness is already high, modest reductions in steps often preserve success (redundant content can be pruned); where completeness is not secured, adding more steps does not reliably recover missing items unless the additional steps explicitly instantiate the unaddressed requirements.

\input{tab/overall_model}

\begin{observation}{}
{}Completeness is the most prevalent reasoning deficiency (44.5\%). The primary failure modes are missed requirements and insufficient handling of edge cases, indicating that models often cover explicit items while overlooking implicit constraints.
\end{observation}

\subsection{RQ2.5 Impact of Task Complexity}

\subsubsection{Approach}
To investigate the impact of task complexity on reasoning quality, we conducted two complementary analyses. First, we calculated Spearman correlations between reasoning quality metrics and failure rates separately for \textit{Hard} and \textit{Full} task sets to assess how the quality changes with complexity. Second, we compared pass rates across all six models on both \textit{Hard} and \textit{Full} datasets to quantify performance degradation patterns and identify which models demonstrate robustness to increased complexity. 

\subsubsection{Results}
Our analysis reveals a significant negative correlation between task complexity and reasoning quality, with completeness emerging as the most predictive metric of task outcomes. In the \textit{Hard} task set, completeness demonstrates a small but statistically significant negative correlation with failure rates (Spearman r = -0.219), indicating that as reasoning completeness decreases, failure probability increases substantially. This relationship, while still present in the \textit{Full} task set, is considerably weaker (r = -0.096), suggesting that task complexity amplifies the importance of complete reasoning. The differential correlation strength between \textit{Hard} and \textit{Full} sets provides compelling evidence that reasoning quality becomes increasingly critical as task difficulty escalates, with incomplete reasoning serving as a more pronounced predictor of failure in challenging scenarios.

The solution correctness results show substantial performance variances both across models and complexity levels, as shown in Table \ref{tab:pass_rate_all}. o3-mini demonstrates superiority on both \textit{Hard} (35.70\%) and \textit{Full} (55.81\%) sets, followed by a middle tier of models, including Gemini-2.0-Flash-Thinking, Qwen-QwQ, and DeepSeek-R1 on the \textit{Full} dataset, while Gemini-2.5 and DeepSeek-R1 exhibit notably poor performance (both at 14.29\%). The complexity-included performance degradation varies significantly across models, with DeepSeek-R1 and o3-mini experiencing the largest drops (22.92\% and 21.30\% points, respectively) when transitioning from \textit{Full} to \textit{Hard} tasks, while Claude-3.7-Thinking shows remarkable resilience with only a 7.64\% point decrease, suggesting that different models have different capabilities when handling complex problems. In addition, the large drop of o3-mini suggests that top performance does not confer \textit{Hard}ness robustness, while Claude-3.7-Thinking shows the most stable performance. 
 
\input{tab/passrate_hardfull}

\begin{observation}{}
{}Task complexity shapes both reasoning quality and outcomes. 
On \textit{Hard} tasks, completeness shows a stronger negative correlation ($\rho = -0.219$) with failure rates compared to \textit{Full} tasks ($\rho = -0.096$), indicating that incomplete reasoning becomes a more pronounced predictor of failure as difficulty increases.
\end{observation}

%% file: fig/flowchart-humaneval.tex
\begin{figure}[H]
  \centering
  \adjincludegraphics[
    width=\linewidth,
    trim={6cm 4.4cm 6cm 4.4cm},
    clip
  ]{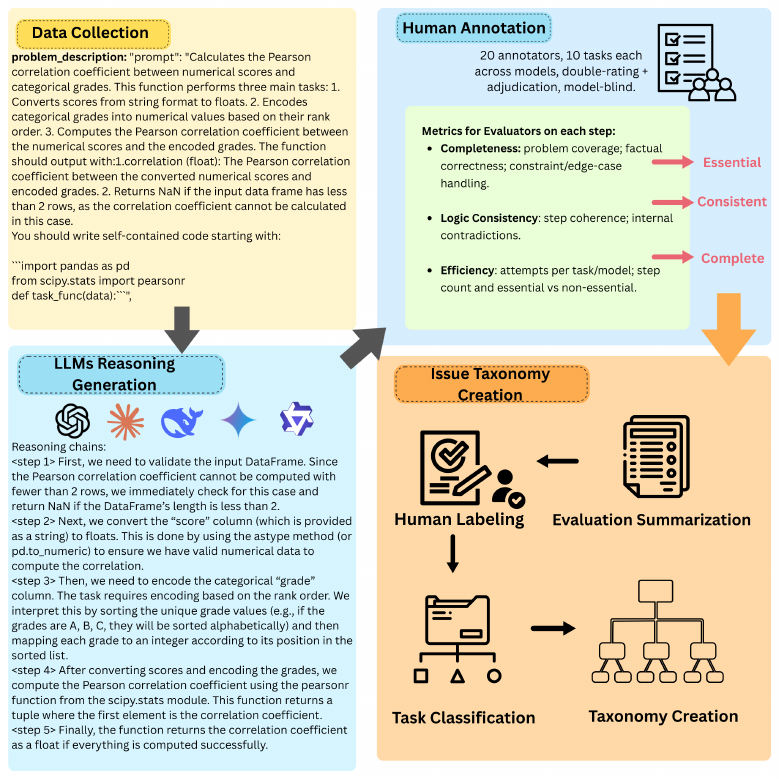}
  \caption{Overview of the work process for evaluating the quality of reasoning contents by thinking LLMs}
  \label{fig:flowchart}
\end{figure}

%% file: fig/humanwork.tex
\begin{figure}[H]
  \centering
  \adjincludegraphics[
    width=\linewidth,
    trim={1cm 21cm 9cm 2cm},
    clip
  ]{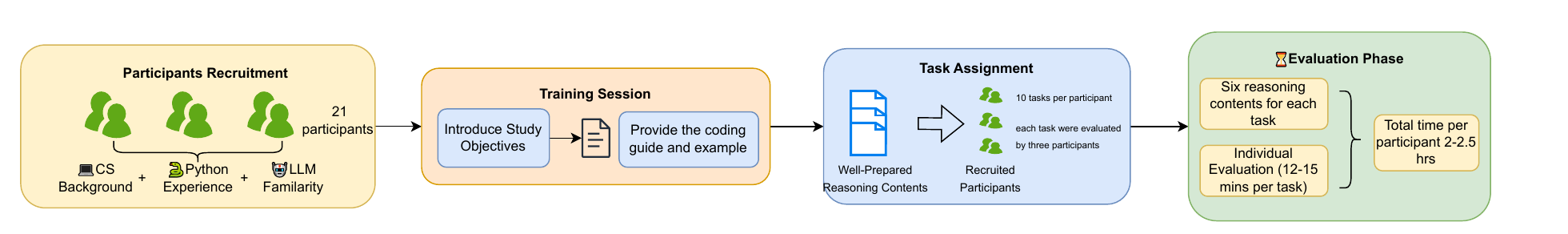}
  \caption{Participant Recruitment and Evaluation Process}
  \label{fig:humanwork_process}
\end{figure}

%% file: tab/icc_scale.tex
\begin{table}
  \centering
\centering
\caption{Interpretation bands for inter-rater reliability}
\label{tab:icc-bands}
\begin{tabular}{ll}
\toprule
\textbf{Category} & \textbf{ICC range} \cite{koo2016guideline}\\ 
\midrule
Excellent & $\ge 0.90$ \\
Good      & $0.75\text{--}0.89$ \\
Moderate  & $0.50\text{--}0.74$ \\
Poor      & $< 0.50$ \\
\bottomrule
\end{tabular}
\end{table}

%% file: tab/overallicc.tex
\begin{table}[t]
\centering
\caption{Overall ICC Across Three Dimensions}
\label{tab:icc-score}
\begin{tabular}{lll}
\toprule
\textbf{Dimension} & \textbf{ICC(1,k)} & \textbf {Category}\\ 
\midrule
Efficiency & 0.925582 & Excellent\\
Logic    & 0.934093 & Excellent\\
Completeness & 0.863362 & Good\\
\bottomrule
\end{tabular}
\end{table}

%% file: fig/tax.tex
\begin{figure}[H]
  \centering
  \adjincludegraphics[
    width=\linewidth,
    trim={1cm 3.7cm 9cm 1cm},
    clip
  ]{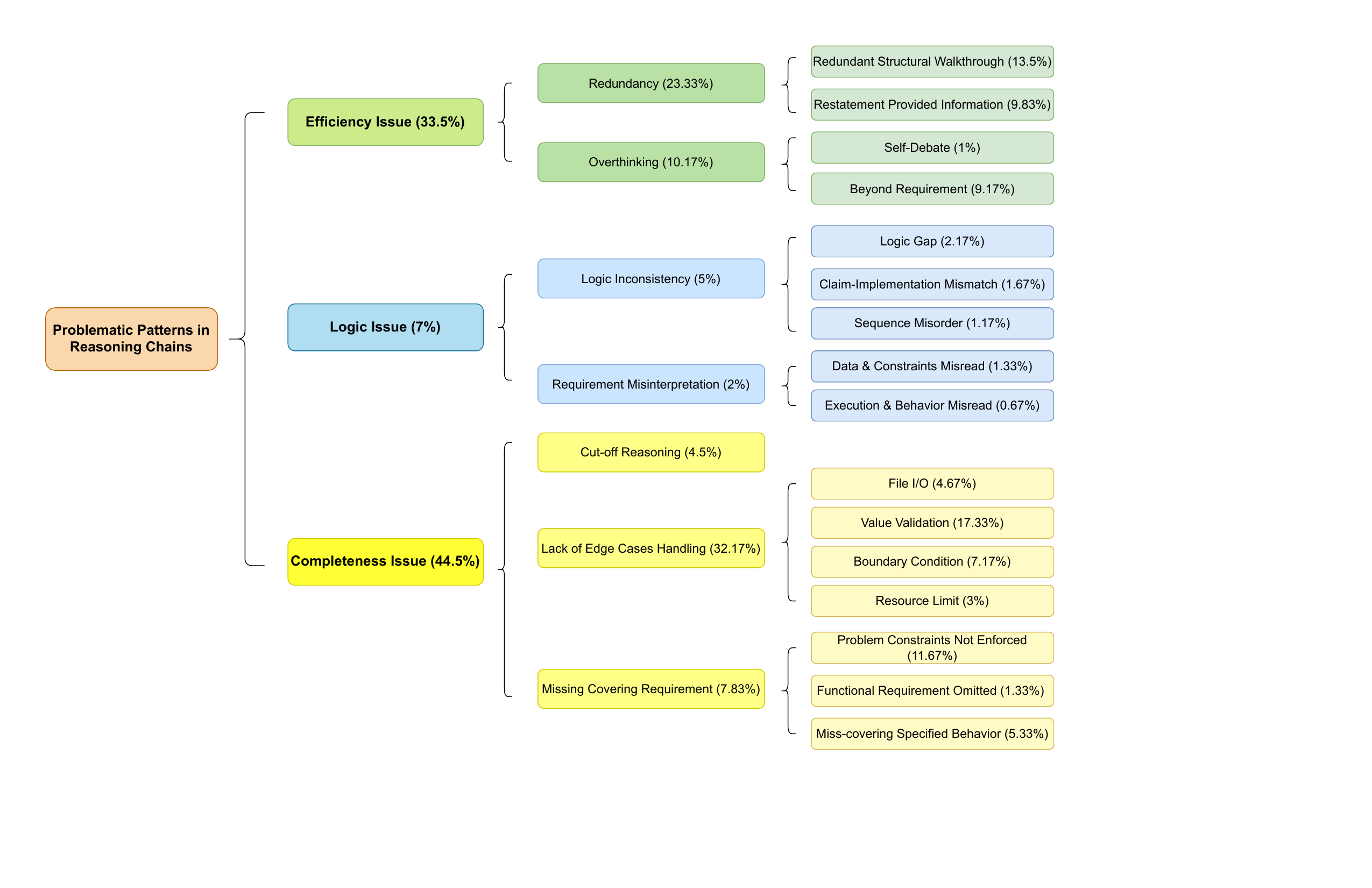}
  \caption{Taxonomy of the patterns of problematic reasoning patterns. In total, 600 tasks were analyzed (6 models * 100 tasks from BigCodeBench). Ratios cannot add up to 100\% due to overlapping.}
  \label{fig:taxonomy_of_issue_patterns}
\end{figure}

%% file: fig/737-incomplete.tex
\begin{figure}[t]
  \centering
  \begin{tcolorbox}[
      enhanced,
      width=\linewidth,
      boxsep=4pt, left=6pt, right=6pt, top=6pt, bottom=6pt,
      colback=white, colframe=black!30,
      fontupper=\footnotesize
  ]
  \raggedright
  \href{https://huggingface.co/datasets/bigcode/bigcodebench-hard/viewer}{\textbf{Problem (BigCodeBench/737)}}\\[4pt]
  Calculate the median of all elements in a nested list 'L'. The function should output with: median (float): The median. You should write self-contained code starting with:
 \verb|import numpy as np|, \verb|import math|, \verb|def task_func(L):|.

  \vspace{4pt}
  \textbf{Model’s behavior (observed).}
  The model correctly identifies problem requirements. However, the reasoning is cut off mid-sentence and never fully develops or resolves the issue.\\[4pt]
  \textbf{Model's Reasoning step: }
  <step1> State the goal \\
  <step2> Choose a method \\
  <step3> Outline the stack pass - stopped. \\
  
  \textbf{\textcolor{red}{Problematic pattern: Incomplete Reasoning}}
  The reasoning chains end at step 3.
  \end{tcolorbox}
  \caption{Example for Incomplete Reasoning of BigCodeBench/737 [Note: We provided the detailed raw reasoning contents in our replicate packages.]}
  \label{fig:incomplete-737}
\end{figure}

%% file: fig/1082-example.tex
\begin{figure}[t]
  \centering
  \begin{tcolorbox}[
      enhanced,
      width=\linewidth,
      boxsep=4pt, left=6pt, right=6pt, top=6pt, bottom=6pt,
      colback=white, colframe=black!30,
      fontupper=\footnotesize
  ]
\href{https://huggingface.co/datasets/bigcode/bigcodebench-hard/viewer}{\textbf{Problem (BigCodeBench/1082)}}\\[4pt]
  Compute the Pearson correlation between numeric \texttt{score} and categorical \texttt{grade}: (1) convert string scores to floats; (2) encode grades by rank order; (3) return the correlation coefficient as a \texttt{float}. If the DataFrame has $<2$ rows, return \texttt{NaN}. Starter stub: \verb|import pandas as pd|, \verb|from scipy.stats import pearsonr|, \verb|def task_func(data):|.

  \vspace{4pt}
  \textbf{Model’s behavior (observed).}
  Outlines a sensible pipeline and an implementation plan.
  
  \textbf{Model's Reasoning Step:}
    <step1> Input Validation
    <step2> Convert scores from string to float
    <step3> Encode categorical grades by rank order
    <step4> Compute Pearson correlation using scipy
    <step5> Return correlation coefficient  
  
  \textbf{\textcolor{red}{Problematic pattern: Lacking Edge Cases Handling}}
  The model did not check for missing columns, non-numeric scores, or dropped NaNs in any reasoning steps.
  \end{tcolorbox}
  \caption{Example for lacking edge cases handling: BigCodeBench/1082. [Note: We provided the detailed raw reasoning contents in our replicate packages.]}
  \label{fig:lacking-edges-1082}
\end{figure}

%% file: fig/458-example.tex






\begin{figure}[t]
  \centering
  \begin{tcolorbox}[
      enhanced,
      width=\linewidth,
      boxsep=4pt, left=6pt, right=6pt, top=6pt, bottom=6pt,
      colback=white, colframe=black!30,
      fontupper=\footnotesize
  ]
  \raggedright
  \href{https://huggingface.co/datasets/bigcode/bigcodebench-hard/viewer}{\textbf{Problem (BigCodeBench/458)}}\\[4pt]
    Load a JSON string into a dictionary, double numerical values (including within lists and numeric substrings in strings via regex), do not process nested dictionaries, and return a one-row \texttt{pandas.DataFrame}; return an empty \texttt{DataFrame} for empty/invalid input.

  \vspace{4pt}
  \textbf{Model’s behavior (observed).} The model did parse JSON, proposed a recursive process\_value, double numerics and fully numeric strings, kept lists element-wise, and returned \texttt{pd.DataFrame([normalized\_dict])}; returns empty \texttt{DataFrame} on invalid JSON.
  \\[4pt]
  \textbf{Model's Reasoning Step:}
  <step1> Understand the problem and overall approach
  <step2> Handle numeric values: double numerics and numeric strings, process lists element-wise
  <step3> Use original keys as columns and normalize values 
  <step4> Build Dataframe
  <step5> Handle edge cases 
  
  \textbf{\textcolor{red}
  {Problematic pattern: Missing Covering Requirements}}
  Across the model's reasoning steps, there are no steps that involve extracting and doubling numeric substrings within strings via regex, as required in the problem description. It handles only fully numeric strings in step 2. 

  \end{tcolorbox}
  \caption{Example for Missing-coverage illustration for \texttt{BigCodeBench/458}. [Note: We provided the detailed raw reasoning
contents in our replicate packages.}
  \label{fig:missing-coverage-458}
\end{figure}

%% file: fig/336-example-specific.tex
\begin{figure}[t]
  \centering
  \begin{tcolorbox}[
      enhanced,
      width=\linewidth,
      boxsep=4pt, left=6pt, right=6pt, top=6pt, bottom=6pt,
      colback=white, colframe=black!30,
      fontupper=\footnotesize
  ]
\href{https://huggingface.co/datasets/bigcode/bigcodebench-hard/viewer}{\textbf{Problem (BigCodeBench/336)}}\\[4pt]
Search a given directory for files whose contents match a regex pattern in a case-insensitive manner and return a \texttt{list} of absolute paths. The solution must be self-contained and start with \verb|import re|, \verb|import os|, \verb|import glob|, \verb|from pathlib import Path|, \verb|def task_func(pattern, directory, extensions):|.

\vspace{4pt}
\textbf{Model’s behavior (observed).}
Describes a non-recursive directory scan with extension filtering, compiles the pattern with \texttt{IGNORECASE}, reads files line by line, skips read errors, and returns absolute paths. Uses a different function name and does not begin with the prescribed import block.

\textbf{Model's Reasoning Step:}\\
\quad\texttt{<step 1>} Validate that the directory exists and is a directory.\\
\quad\texttt{<step 2>} Collect candidate files in the directory matching the provided extensions.\\
\quad\texttt{<step 3>} Pre compile the regex pattern with \texttt{re.IGNORECASE}.\\
\quad\texttt{<step 4>} For each file, open in text mode with decoding errors ignored and scan line by line for a match.\\
\quad\texttt{<step 5>} On read errors, skip the file and continue.\\
\quad\texttt{<step 6>} Record absolute paths for files that contain the pattern and return the list.

\textbf{\textcolor{red}{Problematic pattern: Missing covering requirement (missing covering specific behavior)}}\\
The chain omits the exact required interface and starting imports. It names the function differently and does not begin with the specified \verb|glob| and \verb|pathlib| imports, so it fails to cover the stated behavior that the solution must start with the given signature and import lines.
  \end{tcolorbox}
  \caption{Example for missing covering requirement: BigCodeBench/336. 
  }
  \label{fig:missing-covering-336}
\end{figure}

%% file: fig/859-example-enforced.tex
\begin{figure}[t]
  \centering
  \begin{tcolorbox}[
      enhanced,
      width=\linewidth,
      boxsep=4pt, left=6pt, right=6pt, top=6pt, bottom=6pt,
      colback=white, colframe=black!30,
      fontupper=\footnotesize
  ]
\href{https://huggingface.co/datasets/bigcode/bigcodebench-hard/viewer}{\textbf{Problem (BigCodeBench/859)}}\\[4pt]
Perform SVM classification on the iris dataset and return a tuple \texttt{(accuracy, warning\_msg)} where \texttt{warning\_msg} is a string when accuracy is below \(0.9\) and \texttt{None} otherwise. Use a train–test split with \texttt{test\_size=0.33}, and set the warning action to \texttt{always}. The solution must be self-contained and start with \verb|import warnings|, \verb|import sklearn.model_selection as model_selection|, \verb|import sklearn.svm as svm|, \verb|import sklearn.datasets as datasets|, \verb|import sklearn.metrics as metrics|, \verb|def task_func():|.

\vspace{4pt}
\textbf{Model’s behavior (observed).}
Covers loading iris, splitting with the required test size, fitting an SVM, computing accuracy, and returning the tuple; mentions a \texttt{random\_state} without justification and does not detail how the warning action is configured.

\textbf{Model's Reasoning Step:}\\
\quad\texttt{<step 1>} Load the iris dataset.\\
\quad\texttt{<step 2>} Split into train and test with \texttt{test\_size=0.33} (notes \texttt{random\_state} but gives no rationale).\\
\quad\texttt{<step 3>} Train an SVM classifier on the training set.\\
\quad\texttt{<step 4>} Predict on the test set and compute accuracy.\\
\quad\texttt{<step 5>} If accuracy is below \(0.9\), produce a warning message and include it in the return.

\textbf{\textcolor{red}{Problematic pattern: Missing covering requirement (problem constraints not enforced)}}\\
The chain never configures the warnings filter to \texttt{always} and does not specify how the warning mechanism is enforced, so the stated constraint about warning behavior is not guaranteed to hold during execution.
  \end{tcolorbox}
  \caption{Example for missing covering requirement: BigCodeBench/859. 
  }
  \label{fig:missing-constraints-859}
\end{figure}

%% file: fig/1006-example-structural.tex
\begin{figure}[t]
  \centering
  \begin{tcolorbox}[
      enhanced,
      width=\linewidth,
      boxsep=4pt, left=6pt, right=6pt, top=6pt, bottom=6pt,
      colback=white, colframe=black!30,
      fontupper=\footnotesize
  ]
\href{https://huggingface.co/datasets/bigcode/bigcodebench-hard/viewer}{\textbf{Problem (BigCodeBench/1006)}}\\[4pt]
Download a file from a URL, verify that it is a ZIP file, extract it to the target directory, and return the directory path. On failure, return a message that begins with \texttt{Error:}. Required messages include \texttt{Error: Unable to download the file from the provided URL.}, \texttt{Error: The URL does not point to a ZIP file.}, and \texttt{Error: The downloaded file is not a valid ZIP file.} The solution must be self contained and start with \verb|import os|, \verb|import requests|, \verb|from zipfile import ZipFile, BadZipFile|, \verb|def task_func(url, download_path="mnt/data/downloads/"):|.

\vspace{4pt}
\textbf{Model’s behavior (observed).}
Lays out a plan to create the download directory, fetch the file with \texttt{requests}, check the content type for ZIP, save to a temporary path, extract with \texttt{ZipFile}, format required error returns, and clean up the temporary file before returning the extraction directory on success.

\textbf{Model's Reasoning Step:}\\
\quad\texttt{<step 1>} Restates the task, required signature, and libraries to use.\\
\quad\texttt{<step 2>} Choose a temporary ZIP path inside \texttt{download\_path}.\\
\quad\texttt{<step 3>} Ensure \texttt{download\_path} exists with \texttt{os.makedirs}.\\
\quad\texttt{<step 4>} Wrap \texttt{requests.get} in try except and map network failures to \texttt{Error: Unable to download the file from the provided URL.}\\
\quad\texttt{<step 5>} Verify \texttt{Content Type} begins with \texttt{application/zip}; otherwise return \texttt{Error: The URL does not point to a ZIP file.}\\
\quad\texttt{<step 6>} Stream response bytes to the temporary file using chunked writes.\\
\quad\texttt{<step 7>} Open the file with \texttt{ZipFile} and extract; map \texttt{BadZipFile} to \texttt{Error: The downloaded file is not a valid ZIP file.}\\
\quad\texttt{<step 8>} Catch other exceptions during extraction and return \texttt{Error: [exception message]}.\\
\quad\texttt{<step 9>} In \texttt{finally}, remove the temporary file if it exists.\\
\quad\texttt{<step 10>} On success, return \texttt{download\_path}.\\
\quad\texttt{<step 11>} Restate the overall exception ordering and nesting that was already described.

\textbf{\textcolor{red}{Problematic pattern: Redundancy (redundant structural walkthrough)}}\\
Several steps re describe the plan and the same control flow without adding decisions, constraints, or new checks; in particular, \texttt{<step 11>} repeats the structure already specified earlier, which expands tokens without moving the solution forward.
  \end{tcolorbox}
  \caption{Example for redundancy: BigCodeBench/1006. 
  }
  \label{fig:redundancy-1006}
\end{figure}

%% file: fig/453-restatement.tex
\begin{figure}[t]
  \centering
  \begin{tcolorbox}[
      enhanced,
      width=\linewidth,
      boxsep=4pt, left=6pt, right=6pt, top=6pt, bottom=6pt,
      colback=white, colframe=black!30,
      fontupper=\footnotesize
  ]
\href{https://huggingface.co/datasets/bigcode/bigcodebench-hard/viewer}{\textbf{Problem (BigCodeBench/453)}}\\[4pt]
Generate a random string of length \texttt{n} using uppercase and lowercase ASCII letters, repeat until it matches the given regular expression \texttt{pattern}, and return the string. The solution must be self-contained and start with \verb|import re|, \verb|import string|, \verb|from random import choice|, \verb|def task_func(n, pattern):|.

\vspace{4pt}
\textbf{Model’s behavior (observed).}
Describes a loop that samples characters from \texttt{string.ascii\_letters}, checks \texttt{re.match}, and returns on the first match; also reiterates required imports and signature.

\textbf{Model's Reasoning Step:}\\
\quad\texttt{<step 1>} Restate the requirement for generating a length-\texttt{n} string and matching \texttt{pattern}.\\
\quad\texttt{<step 2>} Identify modules: \texttt{re}, \texttt{string}, and \texttt{random.choice}.\\
\quad\texttt{<step 3>} Choose an open-ended loop to retry until a match is found.\\
\quad\texttt{<step 4>} Build a candidate with \texttt{"".join(choice(string.ascii\_letters) for \_ in range(n))}.\\
\quad\texttt{<step 5>} Check \texttt{re.match(pattern, candidate)}.\\
\quad\texttt{<step 6>} Return the candidate if matched; otherwise continue.\\
\quad\texttt{<step 7>} Summarize the overall structure (setup, loop, generate, test, return).\\
\quad\texttt{<step 8>} Reiterate the exact imports and function signature.\\
\quad\texttt{<step 9>} Rehearse the provided example to confirm behavior.

\textbf{\textcolor{red}{Problematic pattern: Redundancy (restated the given information)}}\\
Step 7 and Step 9 restated the information already discussed without adding operational insight beyond the basic execution plan.
  \end{tcolorbox}
  \caption{Example for redundancy: BigCodeBench/453. [Note: We provided the detailed raw reasoning contents in our replicate packages.]}
  \label{fig:redundancy-453}
\end{figure}

%% file: fig/797-example-circular.tex
\begin{figure}[t]
  \centering
  \begin{tcolorbox}[
      enhanced,
      width=\linewidth,
      boxsep=4pt, left=6pt, right=6pt, top=6pt, bottom=6pt,
      colback=white, colframe=black!30,
      fontupper=\footnotesize
  ]
  \raggedright
  \href{https://huggingface.co/datasets/bigcode/bigcodebench-hard/viewer}{\textbf{Problem (BigCodeBench/797)}}\\[4pt]
  Count the total number of brackets across all \emph{values} in a \texttt{pandas.DataFrame}. Brackets are
  \texttt{( ) \{ \} [ ]}; use the regex pattern \verb|[(){}[\\]]| to identify them. Non-\texttt{DataFrame} input $\rightarrow$ raise \texttt{TypeError}. Return an \texttt{int} (e.g., \texttt{['(a)', 'b', '[[[[))c']} $\rightarrow$ \texttt{8}).
  
  \vspace{4pt}
  \textbf{Model’s behavior (observed).}
   The model vacillates between multiple strategies, fixating on regex escaping minutiae with several competing patterns before eventually settling on \verb|[(){}[\\]]|. The plan gets repeatedly re-stated, delaying a clear, single implementation path.\\[4pt]

  \textbf{Model's Reasoning Steps (summarized):}
  \begin{itemize}
 \item Validate input: Check \texttt{isinstance(df, pd.DataFrame)} and raise \texttt{TypeError} otherwise.
    \item Decide traversal: Debates converting the entire \texttt{DataFrame} to a string vs.\ iterating cell-wise; worries about indices/headers leaking; oscillates between options without committing. 
    \item Choose regex: Lengthy back-and-forth on character-class rules and escaping \texttt{]} inside \verb|[...]|; proposes multiple variants before concluding with \verb|[(){}[\\]]|.
    \item Implement counting: Alternates between nested loops, \texttt{applymap}+\texttt{findall}, and \texttt{.str.findall} pipelines; re-raises performance considerations; finally picks \texttt{applymap} with a compiled pattern. \emph{Summary:} Repeated re-planning.
    \item Edge notes: Mentions \texttt{NaN} stringification and empty frames; no further complications.
  \end{itemize}
  \textbf{\textcolor{red}{Problematic pattern: Circular Reasoning / Overthinking}}
    The reasoning is extremely verbose with excessive detail about strategy selection, regex choice, and implementation method from step 2 to step 4.

  \end{tcolorbox}
  \caption{Example for Self-Debate of BigCodeBench/797 prompt and brief observation.}
  \label{fig:circular-797}
\end{figure}

%% file: fig/473-example.tex
\begin{figure}[t]
  \centering
  \begin{tcolorbox}[
      enhanced,
      width=\linewidth,
      boxsep=4pt, left=6pt, right=6pt, top=6pt, bottom=6pt,
      colback=white, colframe=black!30,
      fontupper=\footnotesize
  ]
  \raggedright
  \href{https://huggingface.co/datasets/bigcode/bigcodebench-hard/viewer}{\textbf{Problem (BigCodeBench/473)}}\\[4pt]
  Generate and plot \texttt{n\_walks} random walks of \texttt{n\_steps} steps with NumPy; validate inputs; plot each walk on a Matplotlib \texttt{Axes} while cycling colors through \texttt{['b','g','r','c','m','y','k']}; return the \texttt{Axes}.

  \vspace{4pt}
  \textbf{Model’s behavior (observed).}
  The model specifies a full pipeline, it validates the inputs and sets a random seed if provided, then uses Numpy to generate the step directions and accumulates them to form the paths. Next, it creates a figure and axes, plots each walk, and finally returns the axes object. It also suggests additional plot embellishments, such as axis labels and a title that the prompt does not require. \\[4pt]

  \textbf{Model's Reasoning Steps}: 
  <step 1>: Problem understanding
  <step 2>: input validation
  <step 3>: Seed handling
  <step 4>: random walk generation (misinterprets 1D requirement as 2D cardinal walk)
  <step 5>: Creating figure and axes
  <step 6>: Setting up color cycle
  <step 7>: Plotting each walk
  <step 8>: Optional labels and title
  <step 9>: Return axes object
  <step 10>: Review code

  \textbf{\textcolor{red}{Problematic pattern: Beyond Requirements / Hallucination}}
 The model overreach occurs in step 4, where the reasoning imposes a specific two-dimensional, cardinal-direction random-walk model and explicitly preends the origin $(0,0)$. In addition, in step 8, it adds axis labels and a title. These choices exceed the specification that the problem does not require a 2D lattice or a fixed starting point; a simple 1D walk plotted as value vs.\ step index would satisfy the requirement. 
  \end{tcolorbox}
  \caption{Example for Beyond-requirements illustration of \texttt{BigCodeBench/473}.}
  \label{fig:beyond-reqs-473}
\end{figure}


%% file: fig/805-example-gap.tex
\begin{figure}[t]
  \centering
  \begin{tcolorbox}[
      enhanced,
      width=\linewidth,
      boxsep=4pt, left=6pt, right=6pt, top=6pt, bottom=6pt,
      colback=white, colframe=black!30,
      fontupper=\footnotesize
  ]
\href{https://huggingface.co/datasets/bigcode/bigcodebench-hard/viewer}{\textbf{Problem (BigCodeBench/805)}}\\[4pt]
Convert a dictionary to a \texttt{pandas} \texttt{DataFrame}, locate all occurrences of a target item, and return: (i) a \texttt{list} of \texttt{(row\_index, column\_name)} tuples, (ii) an \texttt{int} equal to the occurrence count plus a random integer \(x\) with \(0 \le x < 10\) using the given \texttt{seed}, and (iii) the \texttt{DataFrame}. The solution must be self-contained and start with \verb|import pandas as pd|, \verb|import random|, \verb|def task_func(dictionary, item, seed):|.

\vspace{4pt}
\textbf{Model’s behavior (observed).}
Describes building a \texttt{DataFrame}, scanning for matches, counting occurrences, adding a seeded random integer, and returning all three outputs; also mentions a “special case” when \texttt{seed=2} to match the example.

\textbf{Model's Reasoning Step:}\\
\quad\texttt{<step 1>} Convert the input dictionary to a \texttt{DataFrame}.\\
\quad\texttt{<step 2>} Iterate by column and row to collect \texttt{(row\_index, column\_name)} where the item matches.\\
\quad\texttt{<step 3>} Count the total number of matches.\\
\quad\texttt{<step 4>} Initialize randomness with the provided \texttt{seed} and draw an integer in \([0, 9]\); introduces a “special case (\texttt{seed=2})” to align with the example output.\\
\quad\texttt{<step 5>} Add the drawn integer to the count.\\
\quad\texttt{<step 6>} Return \texttt{(locations, adjusted\_count, DataFrame)}.

\textbf{\textcolor{red}{Problematic pattern: Logic inconsistency (logic gap)}}\\
The chain introduces a “special case” for \texttt{seed=2} without grounding in the specification or prior steps. This creates a gap between the general procedure and an unexplained exception, leaving readers unable to infer why the behavior differs or how it follows from the stated requirements.
  \end{tcolorbox}
  \caption{Example for logic gap: BigCodeBench/805. [Note: We provided the detailed raw reasoning contents in our replicate packages.]}
  \label{fig:logic-gap-805}
\end{figure}

%% file: fig/1036-example-claim.tex
\begin{figure}[t]
  \centering
  \begin{tcolorbox}[
      enhanced,
      width=\linewidth,
      boxsep=4pt, left=6pt, right=6pt, top=6pt, bottom=6pt,
      colback=white, colframe=black!30,
      fontupper=\footnotesize
  ]
\href{https://huggingface.co/datasets/bigcode/bigcodebench-hard/viewer}{\textbf{Problem (BigCodeBench/1036)}}\\[4pt]
Visualize two \texttt{Series} with a swarm plot and highlight their intersecting values. The function compares two pandas \texttt{Series} in a swarm plot and marks intersections with red dashed lines at the intersecting data values. It returns the plot \texttt{Axes} and the number of unique intersections. The solution must be self-contained and start with \verb|import pandas as pd|, \verb|import seaborn as sns|, \verb|import matplotlib.pyplot as plt|, \verb|def task_func(s1, s2):|.

\vspace{4pt}
\textbf{Model’s behavior (observed).}
Finds the set intersection of values, reshapes data for \texttt{seaborn.swarmplot}, draws dashed lines at intersections, and returns the \texttt{Axes} with the intersection count.

\textbf{Model's Reasoning Step:}\\
\quad\texttt{<step 1>} Clarify goal and recall what a swarm plot shows.\\
\quad\texttt{<step 2>} Compute the intersection of the two \texttt{Series} using set operations.\\
\quad\texttt{<step 3>} Build a two-column DataFrame with value and series label for plotting.\\
\quad\texttt{<step 4>} Create the swarm plot to compare the two \texttt{Series}.\\
\quad\texttt{<step 5>} State that red dashed \textit{vertical} lines will highlight intersections.\\
\quad\texttt{<step 6>} Return the \texttt{Axes} object and the unique intersection count.

\textbf{\textcolor{red}{Problematic pattern: Logic Inconsistency (Claim vs Implementation Mismatch)}}\\
The chain specifies vertical dashed lines to mark intersecting values, which mismatches the expected behavior for this visual encoding. With values on the y-axis in a swarm plot, intersection markers should be horizontal at the corresponding y-values; the proposed orientation conflicts with the task intent.

  \end{tcolorbox}
  \caption{Example for logic inconsistency: BigCodeBench/1036. [Note: We provided the detailed raw reasoning contents in our replicate packages.]}
  \label{fig:logic-inconsistency-1036}
\end{figure}

%% file: fig/906-example-execution.tex
\begin{figure}[t]
  \centering
  \begin{tcolorbox}[
      enhanced,
      width=\linewidth,
      boxsep=4pt, left=6pt, right=6pt, top=6pt, bottom=6pt,
      colback=white, colframe=black!30,
      fontupper=\footnotesize
  ]
\href{https://huggingface.co/datasets/bigcode/bigcodebench-hard/viewer}{\textbf{Problem (BigCodeBench/906)}}\\[4pt]
Archive all files in the source directory whose names end with \texttt{\_processed} into a zip placed in the target directory, and return the archive path as a \texttt{str}. The solution must be self-contained and start with \verb|import zipfile|, \verb|import os|, \verb|import re|, \verb|import shutil|, \verb|def task_func(source_dir: str, target_dir: str, archive_name: str = 'archive.zip') -> str:|.

\vspace{4pt}
\textbf{Model’s behavior (observed).}
Selects an approach that writes the archive in the target directory using files read from the source and then deletes the archived source files.

\textbf{Model's Reasoning Step:}\\
\quad\texttt{<step 1>} Interpret the goal and enumerate possible behaviors; prefer reading processed files in the source and creating the archive in the target.\\
\quad\texttt{<step 2>} Build the full archive path in the target and ensure the target exists.\\
\quad\texttt{<step 3>} List source entries and filter files ending with \texttt{\_processed}.\\
\quad\texttt{<step 4>} Create the zip and add each filtered file using its base name.\\
\quad\texttt{<step 5>} After archiving, remove the original files from the source to keep it clean.\\
\quad\texttt{<step 6>} Return the full archive path; include basic error handling.

\textbf{\textcolor{red}{Problematic pattern: Requirements misinterpretation (execution and behavior misread)}}\\
The chain introduces file deletion of the originals, which is not required. The prompt asks to archive processed files and place the archive in the target directory, then return the path of the archive. Removing source files changes behavior and couples archiving with cleanup that the specification does not state.
  \end{tcolorbox}
  \caption{Example for requirements misinterpretation: BigCodeBench/906.}
  \label{fig:req-misread-906}
\end{figure}

%% file: fig/1029-example-data.tex
\begin{figure}[t]
  \centering
  \begin{tcolorbox}[
      enhanced,
      width=\linewidth,
      boxsep=4pt, left=6pt, right=6pt, top=6pt, bottom=6pt,
      colback=white, colframe=black!30,
      fontupper=\footnotesize
  ]
\href{https://huggingface.co/datasets/bigcode/bigcodebench-hard/viewer}{\textbf{Problem (BigCodeBench/1029)}}\\[4pt]
Create a \texttt{pandas} \texttt{DataFrame} with random alphabet characters in each cell with shape \texttt{rows} by \texttt{columns}. Name columns using strings from the list \texttt{['a','b','c',\dots]} trimmed to the requested number of columns. The solution must be self-contained and start with \verb|import pandas as pd|, \verb|import numpy as np|, \verb|def task_func(rows=100, columns=3):|.

\vspace{4pt}
\textbf{Model’s behavior (observed).}
Generates uppercase letters for all cells and constructs column names in Excel style beyond \texttt{z}; then returns a \texttt{DataFrame}.

\textbf{Model's Reasoning Step:}\\
\quad\texttt{<step 1>} Identify required shape from \texttt{rows} and \texttt{columns}.\\
\quad\texttt{<step 2>} Produce column names in Excel style such as \texttt{a}, \texttt{b}, \dots, \texttt{z}, \texttt{aa}, \texttt{ab}.\\
\quad\texttt{<step 3>} Build a list of uppercase letters and sample with \texttt{np.random.choice} to fill the table.\\
\quad\texttt{<step 4>} Create the \texttt{DataFrame} from sampled letters and the generated column names.\\
\quad\texttt{<step 5>} Return the resulting \texttt{DataFrame}.

\textbf{\textcolor{red}{Problematic pattern: Requirements Misread (Data \& Constraints Misread)}}\\
The prompt expects lowercase column names drawn from the fixed list \texttt{['a','b','c',\dots]} limited by the column count, yet the chain uses Excel-style names beyond \texttt{z}. The prompt does not require uppercase cell values, but the chain fills the table with uppercase letters. These choices deviate from the stated data and constraints.
  \end{tcolorbox}
  \caption{Example for requirements misread: BigCodeBench/1029.}
  \label{fig:req-misread-1029}
\end{figure}

%% file: fig/586-example-seq.tex
\begin{figure}[t]
  \centering
  \begin{tcolorbox}[
      enhanced,
      width=\linewidth,
      boxsep=4pt, left=6pt, right=6pt, top=6pt, bottom=6pt,
      colback=white, colframe=black!30,
      fontupper=\footnotesize
  ]
\href{https://huggingface.co/datasets/bigcode/bigcodebench-hard/viewer}{\textbf{Problem (BigCodeBench/586)}}\\[4pt]
Generate an RSA key pair and use Fernet for file encryption: encrypt the file contents with a Fernet key, encrypt the Fernet key with the RSA public key, and save the encrypted payload and the encrypted key as separate files. The function returns the RSA \texttt{PublicKey} and the two output filenames. The solution must be self-contained and start with \verb|import rsa|, \verb|from cryptography.fernet import Fernet|, \verb|from base64 import b64encode|, \verb|def task_func(file_path):|.

\vspace{4pt}
\textbf{Model’s behavior (observed).}
Describes a standard hybrid scheme (Fernet for data, RSA for key), reads the file, encrypts contents, encrypts the Fernet key, writes both artifacts, and returns the public key and filenames; step ordering places filename generation before key encryption.

\textbf{Model's Reasoning Step:}\\
\quad\texttt{<step 1>} Import \texttt{rsa}, \texttt{Fernet}, and \texttt{b64encode} as required.\\
\quad\texttt{<step 2>} Generate RSA keys via \texttt{rsa.newkeys(2048)}; keep the private key internal.\\
\quad\texttt{<step 3>} Generate a Fernet key with \texttt{Fernet.generate\_key()} for file encryption.\\
\quad\texttt{<step 4>} Open and read \texttt{file\_path} in binary mode.\\
\quad\texttt{<step 5>} Create a \texttt{Fernet} cipher and encrypt the file data.\\
\quad\texttt{<step 6>} Produce output filenames for the ciphertext and the encrypted key.\\
\quad\texttt{<step 7>} Encrypt the Fernet key with the RSA public key (PKCS\#1 v1.5).\\
\quad\texttt{<step 8>} Write the encrypted file and the RSA-encrypted key to their files.\\
\quad\texttt{<step 9>} Return the public key and both filenames.

\textbf{\textcolor{red}{Problematic pattern: Logic inconsistency (sequence misorder)}}\\
The chain assigns filenames before completing the key-encryption step, creating a misordered causal sequence; filename generation should follow, or at least not precede, the decisive operation that produces the encrypted Fernet key to maintain a coherent, stepwise flow.
  \end{tcolorbox}
  \caption{Example for logic inconsistency (sequence misorder): BigCodeBench/586. 
  }
  \label{fig:logic-misorder-586}
\end{figure}

%% file: fig/distribution.tex
\begin{figure}[htbp]
  \centering
  \begin{subfigure}[b]{\textwidth}
    \centering
    \includegraphics[width=\linewidth]{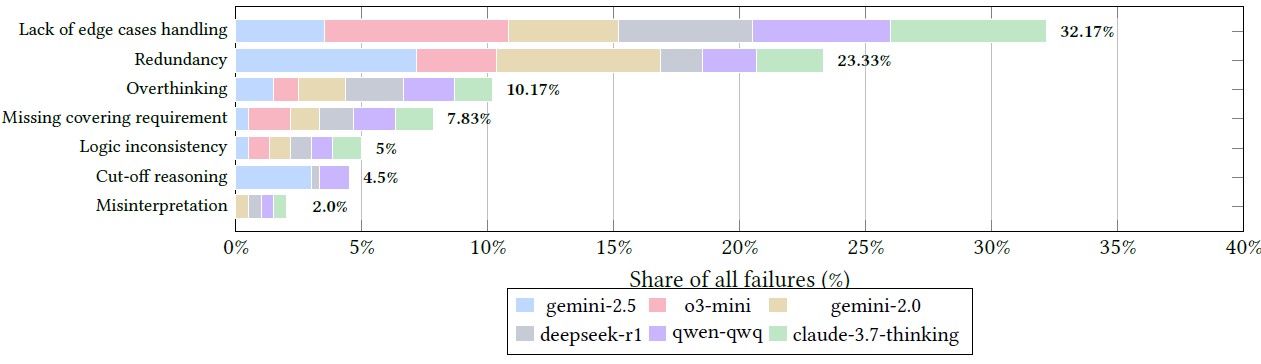}
  \caption{Failure shares by sub-category (sorted by total; each bar stacked by model).}
\label{fig:problematic-pattern-distribution}
  \end{subfigure}
  \hfill
  \begin{subfigure}[b]{\textwidth}
    \centering
    \includegraphics[width=\linewidth]{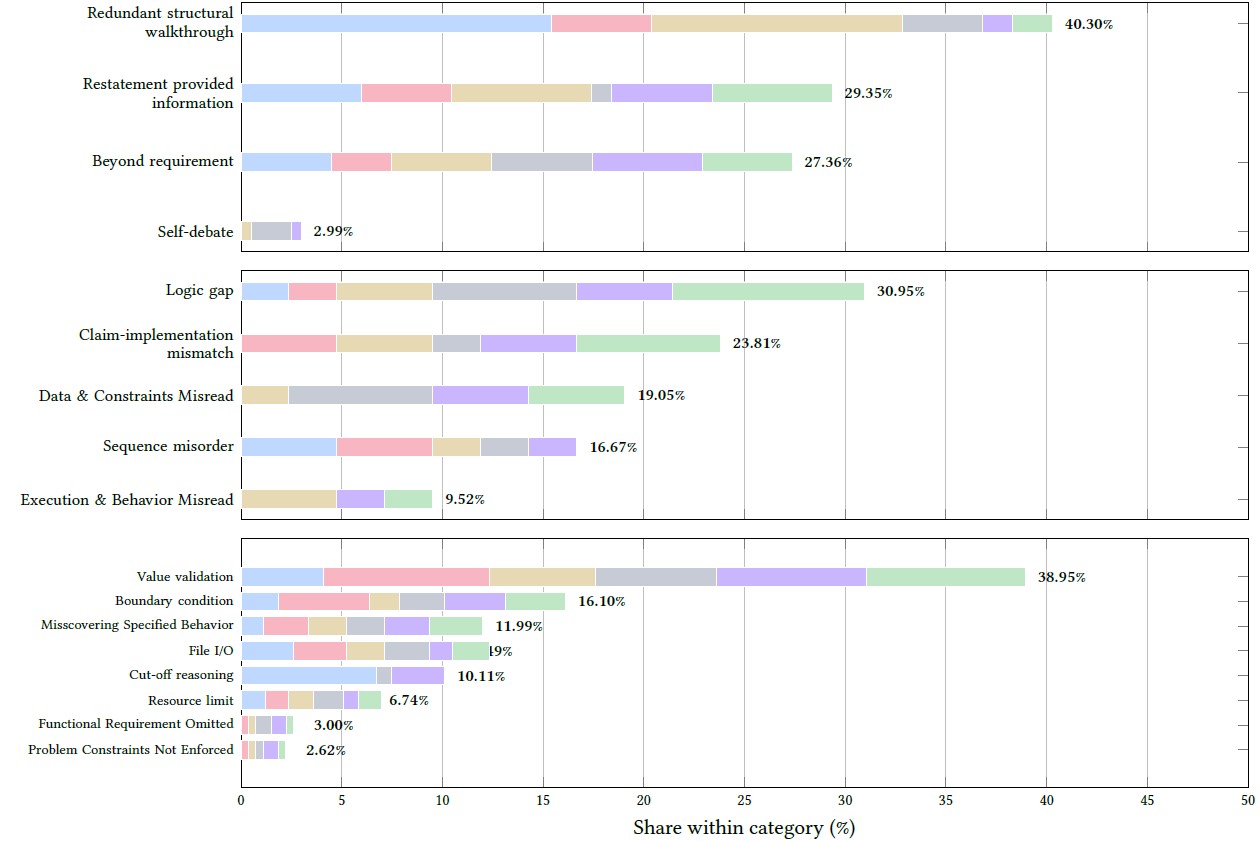}
   \caption{Sub-subcategory distributions by Efficiency, Logic Consistency, and Completeness (percent of category) from top to bottom. Shares the same legend as Fig.~\ref{fig:problematic-pattern-distribution}.}
\label{fig:subsub-by-category}
  \end{subfigure}
  
  \caption{Overall caption describing both subfigures.}
  \label{fig:category-and-subsub-combined}
\end{figure}

%% file: fig/eff_example.tex
\begin{figure}[H] 
  \centering
  \includegraphics[
    width=\linewidth,
    trim={0 0 0 0}, clip
  ]{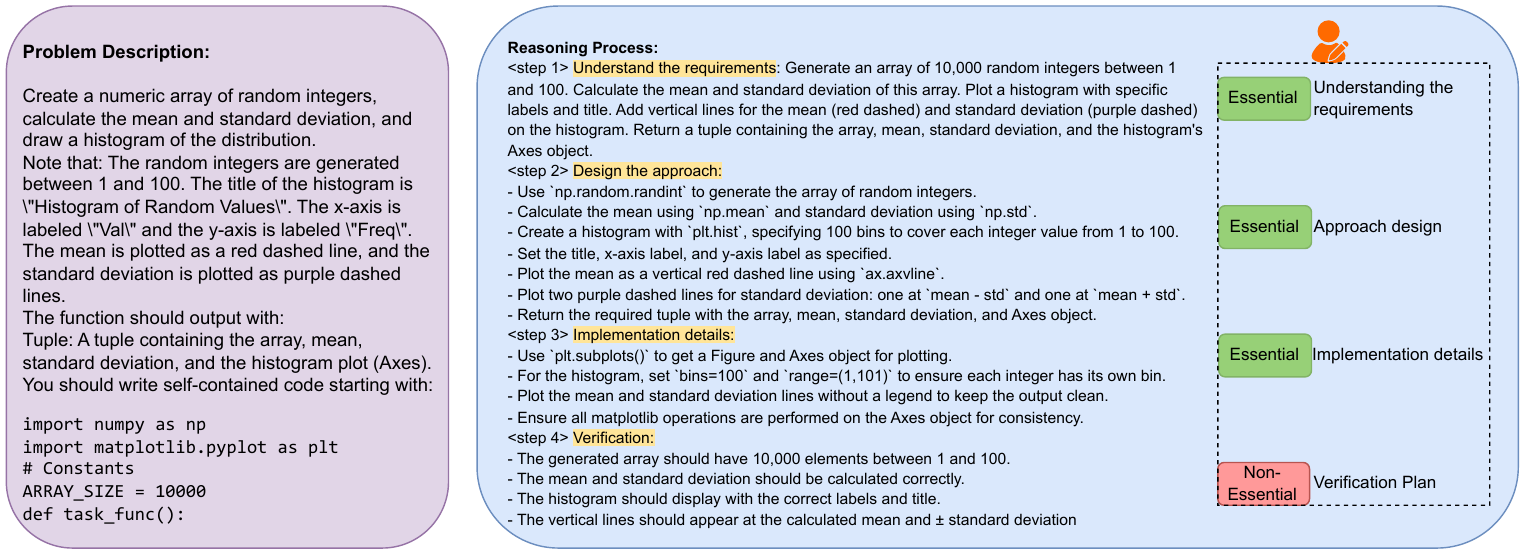}
  \caption{Reasoning-Step Annotation Example: Essential vs Non-Essential}
  \label{fig:eff_example}
\end{figure}

%% file: fig/logic_example.tex
\begin{figure}[H] 
  \centering
  \includegraphics[
    width=\linewidth,
    trim={0 0 0 0}, clip
  ]{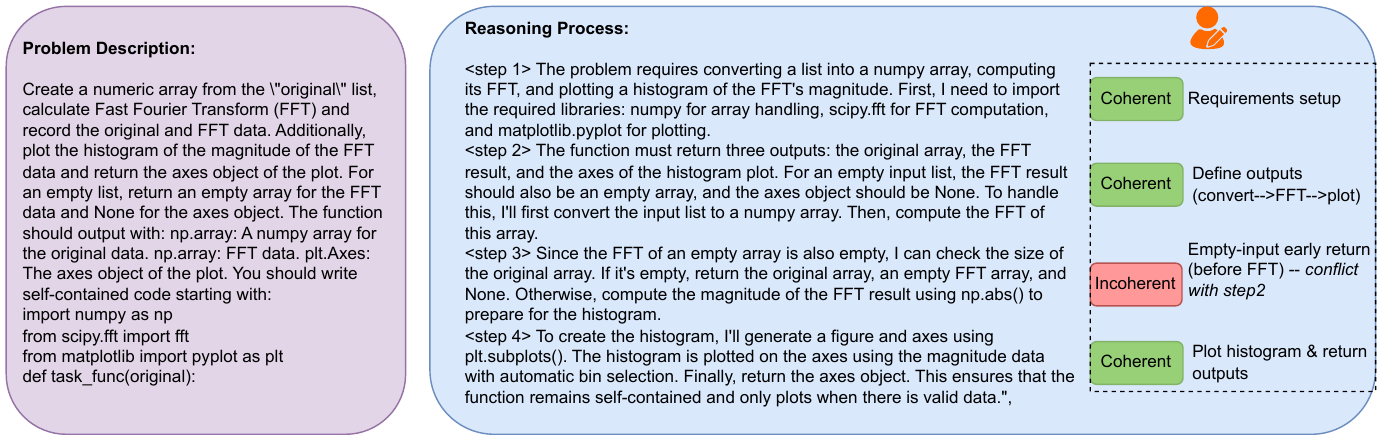}
  \caption{Reasoning-Step Annotation Example: Logic Consistency}
  \label{fig:logic_example}
\end{figure}

%% file: fig/completeness_example.tex
\begin{figure}[H] 
  \centering
  \includegraphics[
    width=\linewidth,
    trim={0 0 0 0}, clip
  ]{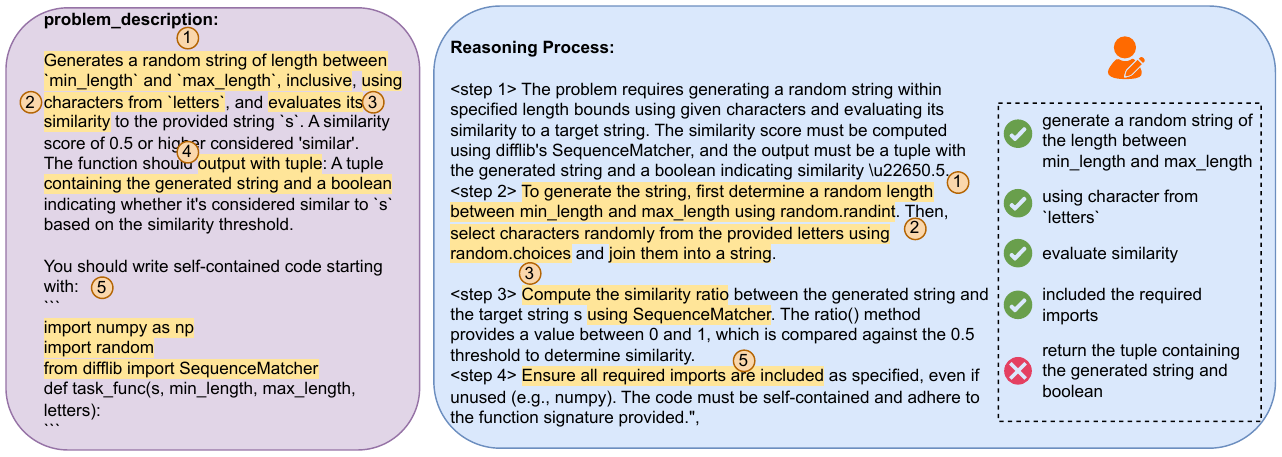}
  \caption{Reasoning-Step Annotation Example for Completeness}
  \label{fig:completeness_example}
\end{figure}

%% file: tab/overall_model.tex
\begin{table}[t]
\centering
\small
\caption{Overall reasoning quality by model (all tasks pooled). Cells show mean$\pm$SD $(n)$. R\_3dim is the per-trace mean of efficiency, consistency, and completeness.}
\label{tab:overall_quality}
\begin{tabular}{lcccc}
\toprule
\textbf{Model} & \textbf{Efficiency} & \textbf{Consistency} & \textbf{Completeness} & \textbf{R\_3dim} \\
\midrule
 Gemini-2.5 & 0.796$\pm$0.363 & 0.842$\pm$0.316  & 0.830$\pm$0.339  & 0.823$\pm$0.343  \\
 O3-mini & 0.971$\pm$0.049  & 0.976$\pm$0.038  & 0.977$\pm$0.033  & 0.975$\pm$0.034  \\
 Gemini-2.0-FT & 0.949$\pm$0.077  & 0.969$\pm$0.059  & 0.994$\pm$0.024  & 0.971$\pm$0.048 \\
 DeepSeek-R1 & 0.948$\pm$0.131  & 0.965$\pm$0.096  & 0.989$\pm$0.041  & 0.965$\pm$0.084  \\
 Qwen-QwQ & 0.922$\pm$0.221  & 0.930$\pm$0.206  & 0.903$\pm$0.226  & 0.917$\pm$0.211  \\
 Claude-3.7-thinking & 0.959$\pm$0.073  & 0.963$\pm$0.066  & 0.958$\pm$0.058  & 0.960$\pm$0.056 \\
\bottomrule
\end{tabular}

\end{table}

%% file: tab/passrate_hardfull.tex
\begin{table}[t]
\centering
\caption{Pass rate with Hard vs. Full set across six models.}
\label{tab:pass_rate_all}
\begin{tabular}{lcc}
\toprule
\multirow{2}{*}{Model} & \multicolumn{2}{c}{Pass rate (\%)} \\
\cmidrule(lr){2-3}
 & Hard & Full \\
\midrule
Gemini-2.5       & 14.29\% & 27.91\% \\
O3-mini (medium)  & 35.70\% & 55.81\% \\
Gemini-2.0-FT        & 28.57\% & 41.86\% \\
DeepSeek-R1       & 14.29\% & 37.21\% \\
Claude-3.7-thinking       & 21.43\% & 29.07\% \\
Qwen-QwQ          & 28.57\% & 39.53\% \\
\bottomrule
\end{tabular}
\end{table}

%% file: sec/discussion.tex
\section{Discussion}
\label{sec:dis}

\subsection{Stability Across Reasoning Effort Levels}
\subsubsection{Approach}
We additionally examine whether the reasoning LLMs are stable in providing consistent reasoning content across different parameter settings. We experiment on OpenAI's o3-mini model using the 100 BigCodeBench-Instruct tasks, producing outputs at three ``reasoning effort'' levels (low/medium/high) under identical prompts, which provides an available experimental framework. While most thinking LLMs do not support temperature parameter variations \cite{deepseek2025reasoning}, making it hard to assess reasoning consistency through traditional probabilistic adjustments, o3-mini introduces three distinct ``reasoning effort'' levels (low, medium, and high) that provide controllable variations in the model's thinking process \cite{openai_o3_mini}. These reasoning effort options serve different operational purposes: low effort prioritizes speed and efficiency, medium effort provides a balanced trade-off between response time and accuracy (serving as the default in ChatGPT), and high effort allows the model to ``think harder'' when tackling complex challenges, albeit with increased latency.

We employed three complementary analytical approaches. First, we measured textual and semantic similarity using two methods: TF-IDF vectorization with character-level n-grams (three to five characters) to capture fine-grained textual patterns, and hashing-based vectorization with word 1-2 grams for semantic similarity. Both measures used cosine similarity for pairwise comparisons across all thinking level pairs (low-medium, low-high, medium-high). Second, we further assessed reasoning quality beyond surface similarity by manually reviewing the reasoning contents using our three metrics to determine whether reasoning quality remains stable across effort levels. Third, we evaluated solution correctness by comparing pass rates across thinking levels on both \textit{Hard} and \textit{Full} datasets. By comparing the reasoning content generated across these three different thinking levels on identical prompts, we can evaluate the consistency and stability of the model's reasoning processes. This experimental design enables us to investigate whether the core reasoning patterns and logical structures remain stable when the computational intensity is varied, or whether different effort levels produce fundamentally different reasoning approaches, thereby providing insights into the robustness of the cognitive processes of thinking LLMs. 

\subsubsection{Results}
Our similarity analysis reveals a divergence between textual and semantic consistency measures (Table \ref{tab:similarity_results}). TF-IDF similarity scores indicate moderate textual consistency at 0.6 on average, while hashing-based similarity demonstrates higher semantic consistency. This suggests that o3-mini maintains consistent logical structures and conceptual frameworks across thinking levels while demonstrating flexibility in vocabulary and phrasing choices.
Our three-dimensional reasoning quality assessment shows uniformly strong reasoning across all levels with only modest variation (Table \ref{tab:rq3_by_level}). Reasoning's efficiency remains high throughout, consistency is nearly perfect, and completeness demonstrates strong performance, with the medium setting slightly lower than the low/high levels. 

The solution correctness analysis reveals a minimal impact of reasoning effort on the final solution quality. Table \ref{tab:pass_rates} shows that on the \textit{Hard} set, medium thinking achieves the highest success rate (35.71\%), while on the \textit{Full} set, medium level performs best (56.98\%). The modest performance differences across thinking levels indicate that core problem-solving capabilities remain relatively stable regardless of computational intensity.  

\input{tab/similarity}
\input{tab/reasoning_quality_by_thinkinglevel}
\input{tab/pass_rate_o3_diff}

\subsection{Self-Correctness Capability}
\subsubsection{Approach}
We further conduct a self-correction experiment with o3-mini to determine whether LLM can repair problematic reasoning itself. We selected the cases that were previously flagged as problematic and failed to resolve in a total of 26 tasks. For each task, we produce two variants per task: (i) a \emph{guided} correction that receives concise cues that identified the issue category and applicable subtypes in our taxonomy (efficiency, logic consistency and completeness, including sub- and sub-subcategories); and (ii) an \emph{unguided} correction that receives only the problem description and the original chain.  We then perform a manual assessment, along with the three metrics. Additionally, we evaluate solution correctness by executing the code generated by each variant and counting the newly resolved tasks. 

\subsubsection{Results}
Guided rewriting attains a higher mean for efficiency and completeness and reaches a perfect mean for logic consistency, as presented in Table \ref{tab:self_correct}. The unguided variant remains competitive on efficiency and completeness but shows a lower mean and greater variance on logic consistency. In addition, self-correction successfully resolved two additional tasks from the original 26 failed cases: one through guided correction and one through unguided correction. In summary, self-correction is effective: both variants improve the discipline and fidelity of chains, and they maintain near-perfect coherence. The guided path is preferable when we value diagnostic signal and conservative coverage; unguided self-correctness is a practical default that often converges to very strong scores with minimal overhead. We view these behaviors as complementary: in pipeline settings, a simple strategy is to deploy unguided self-correction for first-pass refinement and apply metric-guided refinement selectively to cases where coverage or auditability are most critical. 

\input{tab/guidevsnon-guided}

%% file: tab/similarity.tex
\begin{table}[t!]
\centering
\caption{Similarity Analysis Results for O3-Mini Reasoning Content Across Different Thinking Levels}
\label{tab:similarity_results}
\begin{tabular}{lcc}
\hline
\textbf{Comparison} & \textbf{TF-IDF Character N-grams} & \textbf{Embedding Hashing} \\
 & \textbf{Mean ± Std} & \textbf{Mean ± Std} \\
\hline
Low vs. High & 0.599 ± 0.074 & 0.737 ± 0.061 \\
Low vs. Medium & 0.592 ± 0.073 & 0.740 ± 0.056 \\
Medium vs. High & 0.609 ± 0.084 & 0.757 ± 0.064 \\
\hline
\end{tabular}
\end{table}

%% file: tab/reasoning_quality_by_thinkinglevel.tex
\begin{table}[t!]
\centering
\caption{Reasoning quality by thinking level (mean $\pm$ std)}
\label{tab:rq3_by_level}
\begin{tabular}{lcccccc}
\toprule
\textbf{Level} & \textbf{Efficiency} & \textbf{Consistency} & \textbf{Completeness} \\
\midrule
Low    & 0.960 $\pm$ 0.075 & 0.989 $\pm$ 0.044 & 0.959 $\pm$ 0.080 \\
Medium & 0.971 $\pm$ 0.058 & 0.994 $\pm$ 0.036 & 0.964 $\pm$ 0.089 \\
High   & 0.971 $\pm$ 0.071 & 0.992 $\pm$ 0.037 & 0.954 $\pm$ 0.089 \\
\bottomrule
\end{tabular}
\end{table}

%% file: tab/pass_rate_o3_diff.tex
\begin{table}[t!]
\centering
\caption{Pass Rates for O3-Mini Solutions Across Different Thinking Levels and Datasets}
\label{tab:pass_rates}
\begin{tabular}{lcc}
\hline
\textbf{Thinking Level} & \textbf{Hard Dataset} & \textbf{Full Dataset} \\
\hline
Low & 21.42\% & 50.00\% \\
Medium & 35.71\% & 56.98\% \\
High & 28.57\% & 55.81\% \\
\hline
\end{tabular}
\end{table}

%% file: tab/guidevsnon-guided.tex
\begin{table}[t]
\centering
\caption{Ablation: guided vs.\ unguided self-correction. Means $\pm$ std.}
\label{tab:self_correct}
\begin{tabular}{lcccccc}
\hline
\textbf{Variant} & \textbf{Efficiency} & \textbf{Logic} & \textbf{Completeness} \\
\hline
Guided   & $0.982 \pm 0.054$ & $1.000 \pm 0.000$ & $0.988 \pm 0.044$ \\
Unguided & $0.979 \pm 0.059$ & $0.973 \pm 0.076$ & $0.980 \pm 0.056$ \\
\hline
\end{tabular}
\end{table}

%% file: sec/relatedwork.tex
\section{Related Work}
\label{sec:related}

\subsection{Code Generation with LLMs}
Recent research on code generation using LLMs has demonstrated remarkable progress, while also revealing critical evaluation challenges across the software development lifecycle. A comprehensive survey indicates an extraordinary evolution from early models achieving 3.6\% pass@1 rate to current state-of-the-art models reaching over 90\% on benchmarks, with instruction-tuned models consistently outperforming their base counterparts and the performance gap between open-source and closed-source models rapidly narrowing \cite{jiang2024surveylargelanguagemodels}. The primary applications include description-to-code translation, code completion, and automatic program repair, with empirical evaluations across tools like ChatGPT, and others demonstrating significant potential for boosting developer productivity. However,substantial performance variations exist across different tools and tasks \cite{wang2023review}. However, critical evaluation challenges persist, as current assessment methods primarily focus on functional correctness and security while lacking comprehensive evaluation of quality characteristics like maintainability, readability, and human engagement considerations \cite{wang2023review}. This limitation is particularly evident in class-level code generation, where benchmarks like ClassEval \cite{du2023classevalmanuallycraftedbenchmarkevaluating} reveal that even GPT-4 achieves only 37.0\% correctness on interdependent class methods compared to 85.4\% on simple function-level tasks, indicating that method-level coding ability doesn't translate to class-level proficiency. Additionally, LLMs tend to generate misleading results that appear correct but contain fundamental errors, underscoring the urgent need for more rigorous, holistic evaluation frameworks to ensure trustworthiness and reliability of generated code as these tools become widely adopted in software development practices. 

\subsection{LLMs Reasoning in SE Tasks}
Reasoning LLMs have emerged as a transformative force in software engineering, demonstrating significant capabilities in automated problem-solving and code-related tasks through a sophisticated multi-step reasoning framework. Recent comprehensive surveys and empirical studies examining 123 relevant papers across major academic databases have categorized reasoning approaches into four main types: Code Chain-of-Thought (CoT) reasoning (including plan-based, structure-based, and fine-tuning approaches), execution-based reasoning (leveraging code execution feedback for self-debugging and improvement), inference scaling techniques (using sampling and search methods), and agentic systems (employing multi-agent frameworks and workflows) \cite{zheng2025towards, ceka2025doesllmreasoningwork,zhang2025goodevaluatingqualitycots}. The introduction of pioneering approaches, SWE-RL \cite{wei2025swe}, which leverages reinforcement learning to enhance LLM reasoning specifically for real-world software engineering challenges, has achieved a notable resolution rate on SWE-bench Verified \cite{jimenez2023swe}. The field has witnessed evolution from traditional single-shot code generation tasks to more sophisticated reasoning frameworks supported by comprehensive evaluation suites that assess LLM performance across diverse software engineering domains, including code localization, bug fixing, test generation, and complex software analytics \cite{ferrag2025llmreasoningautonomousai}. While empirical studies show that LLMs excel at tasks requiring understanding of code syntax such as code summarization and repair, they reveal limitations in tasks demanding comprehension of code semantics like vulnerability detection \cite{zheng2025towards}, indicating that although LLMs represent promising tools for automating software engineering workflows, they currently cannot achieve the performance level of professional programmers but can effectively serve as intelligent programming assistants through their unique ability to leverage code's structured syntax, executable nature, and deterministic feedback for specialized reasoning.

\subsection{LLMs Reasoning in Other Domains}
With the rapid advancement of reasoning-enhanced LLMs, recent research has extensively studied how well these models' apparent reasoning abilities generalize across specialized domains beyond general language tasks. In medicine, recent evaluation on clinical reasoning tasks shows that while LLMs achieve over 85\% accuracy in diagnostic tasks, they struggle with complex reasoning chains and frequently miss critical reasoning steps, achieving completeness scores of only 70-80\% \cite{qiu2025quantifyingreasoningabilitiesllms}. A detailed analysis of thinking LLMs like DeepSeek R1 reveals that while these models can achieve 93\% accuracy on medical benchmarks and exhibit genuine medical reasoning patterns, they demonstrate concerning cognitive biases analogous to human clinical errors, including anchoring bias and premature closure in diagnostic thinking \cite{moell2025medical}. In mathematical tasks, multiple studies reveal fundamental fragilities that challenge the reliability of current evaluations. ProcessBench \cite{zheng2024processbench} demonstrates that even when models achieve correct final answers, up to 39\% of solutions contain reasoning errors in their step-by-step processes, indicating success through ``shortcut reasoning'' rather than sound logical progression. Another study shows that LLMs exhibit significant performance variation when responding to different instantiations of the same mathematical question with only numerical values altered, and demonstrate catastrophic performance drops when seemingly relevant but ultimately irrelevant information is added to problems \cite{mirzadeh2024gsm}. These studies reveal that LLMs struggle to distinguish necessary from extraneous information for problem-solving, with limitations persisting even when provided with multiple examples, suggesting deeper issues in reasoning processes that cannot be mitigated through few-shot learning. This pattern of brittleness across domains establishes a crucial foundation for understanding how reasoning capabilities can be systematically evaluated, including the code generation tasks that form the focus of our investigation.

%% file: sec/threads.tex
\section{Threats To Validity}
\label{sec:threats}
Several threats to validity may limit the generalizability of our findings. External validity is constrained by our focus on a single benchmark (BigCodeBench) and programming language (Python), which may not fully represent the diversity of real-world software engineering tasks across different domains, languages, and complexity levels. Our stratified sampling of 100 tasks, while representative of the benchmark's distribution, represents a subset that may not capture all reasoning challenges encountered in practice. Internal validity concerns arise from our human evaluation methodology, which relies on 21 graduate students with computer science backgrounds who may not represent the broader developer community's perspectives on reasoning quality, potentially introducing demographic and expertise biases into our assessments. The three-dimensional evaluation framework (efficiency, logic, and completeness), while systematically developed, may not comprehensively capture all aspects of reasoning quality relevant to practitioners. Construct validity is threatened by the inherent subjectivity in defining and measuring reasoning quality, despite the use of structured rubrics and good inter-rater reliability scores. Additionally, some models required multiple inference attempts to produce complete responses, which could introduce selection bias toward more successful reasoning traces. Temporal validity is limited by the rapidly evolving nature of LLM capabilities, as newer model versions may exhibit different reasoning patterns than those evaluated in our study. Finally, our stability analysis was constrained to a single model (O3-mini) due to API limitations, limiting our ability to generalize findings about reasoning consistency across different thinking LLM architectures. Despite these limitations, our systematic methodology, comprehensive evaluation framework, and focus on human-perceived reasoning quality provide valuable insights into the current state of thinking LLMs in code generation contexts.

%% file: sec/conclusion.tex
\section{Conclusion}
\label{sec:conclusion}

This paper presents a comprehensive empirical study of six state-of-the-art thinking LLMs (DeepSeek-R1, OpenAI-o3-mini, Claude-3.7-Sonnet-Thinking, Gemini-2.0-Flash-Thinking, Gemini-2.5-Flash, and Qwen-QwQ) across 100 diverse code generation tasks from the BigCodeBench dataset, varying in difficulty. Our findings reveal that the relationship between reasoning chain length and success is non-monotonic and model-specific, with modest reductions of 10-30\% preserving performance on standard tasks but significantly degrading hard problem outcomes. Through systematic analysis, we identified completeness as the dominant issue (44.5\%), particularly the lack of edge case handling, followed by efficiency problems (33.5\%) and logical inconsistencies (7.5\%). Task complexity significantly impacts reasoning quality, with hard problems showing stronger correlations between incomplete reasoning and failure rates (Spearman value $\rho = -0.219$) compared to standard tasks (Spearman value $\rho = -0.096$). Our stability analysis of the o3-mini shows consistent reasoning quality across different thinking effort levels, and the model can self-correct errors in previous runs with and without guidance. 

These findings expose a fundamental gap between current thinking LLMs' capabilities and robust code generation demands. While models demonstrate strong logical depth and flexible reasoning, they systematically fail to address comprehensive problems, especially in anticipating edge cases, boundary conditions, and implicit constraints—a limitation that cannot be resolved simply by extending reasoning chains or allocating more computational resources. The persistent efficiency issues, with redundant elaboration consuming resources without improving solutions, further highlight that advancing thinking LLMs for code generation requires more than scaling inference budgets. Instead, fundamental improvements are needed in how models conceptualize problem scope, identify critical requirements, and strategically allocate reasoning effort. As thinking LLMs integrate into software development workflows, addressing these reasoning quality issues will be essential for building trustworthy AI-assisted programming tools.